\documentclass{article}

\usepackage[preprint]{neurips_2022}

\usepackage[utf8]{inputenc} %
\usepackage[T1]{fontenc}    %
\usepackage{hyperref}       %
\usepackage{url}            %
\usepackage{booktabs}       %
\usepackage{amsfonts}       %
\usepackage{nicefrac}       %
\usepackage{microtype}      %
\usepackage{xcolor}         %
\usepackage{cite}
\usepackage{amsmath}
\usepackage{tikz}
\usepackage{subcaption}
\usepackage{longtable}
\usepackage{printlen}  %
\usepackage{dcolumn}
\usepackage{xspace}
\usepackage{pifont}
\usepackage{cleveref}
\newcommand{\cmark}{\ding{51}}%
\newcommand{\xmark}{\ding{55}}%
\definecolor{dukeblue}{rgb}{0.0, 0.0, 0.61}
\newcommand{\ours}{SymFormer\xspace}
\newcommand{\Ours}{SymFormer\xspace}

\newcommand{\eg}{\textit{e.g.}\xspace}
\newcommand{\supp}{\textit{Supp. Mat.}\xspace}

\title{SymFormer: End-to-end symbolic regression using transformer-based architecture}

\newcommand\inst[1]{$^{#1}$}
\author{%
Martin Vastl\inst{1,2} \and
\textbf{Jonáš Kulhánek}\inst{1,3} \and
\textbf{Jiří Kubalík}\inst{1} \and
\textbf{Erik Derner}\inst{1,3} \and
\textbf{Robert Babuška}\inst{1,4}
\AND
{\normalfont\inst{1} Czech Institute of Informatics, Robotics and Cybernetics,}\\ Czech Technical University in Prague\\
\texttt{\{vastlmar,jonas.kulhanek,jiri.kubalik,erik.derner\}@cvut.cz} \and
\inst{2} Faculty of Mathematics and Physics, Charles University \and
\inst{3} Faculty of Electrical Engineering, Czech Technical University in Prague \and
\inst{4} Cognitive Robotics Faculty of 3mE, Delft University of Technology\\
\texttt{r.babuska@tudelft.nl}
}

\begin{document}

\maketitle

\begin{abstract}
Many real-world problems can be naturally described by mathematical formulas. The task of finding formulas from a set of observed inputs and outputs is called symbolic regression. Recently, neural networks have been applied to symbolic regression, among which the transformer-based ones seem to be the most promising. After training the transformer on a large number of formulas (in the order of days), the actual inference, i.e., finding a formula for new, unseen data, is very fast (in the order of seconds). This is considerably faster than state-of-the-art evolutionary methods. The main drawback of transformers is that they generate formulas without numerical constants, which have to be optimized separately, so yielding suboptimal results. We propose a transformer-based approach called SymFormer, which predicts the formula by outputting the individual symbols and the corresponding constants simultaneously. This leads to better performance in terms of fitting the available data. In addition, the constants provided by SymFormer serve as a good starting point for subsequent tuning via gradient descent to further improve the performance. We show on a set of benchmarks that SymFormer outperforms two state-of-the-art methods while having faster inference.

\end{abstract}

\section{Introduction}
Many natural processes and technical systems can be described by mathematical formulas.
Knowing the correct formula would not only provide us with some insight into the process's inner workings, but  it would also allow us to predict how the process will evolve in the future.
Therefore, being able to automatically derive a simple formula that fits the observed data would have a tremendous impact on applications in all areas of science.
The task of finding such a formula from the observed data is called \emph{symbolic regression} (SR) and has already been applied to a variety of real-world problems, \eg, in physics \citep{wadekar22augmenting, konstantin21analytical}, robotics \citep{kubalik19symbolic, hein17interpretable}, or machine learning \citep{wilstrup21symbolic}.

Historically, most of the symbolic regression methods \citep{schmidt09distilling, kubalik20symbolic} were tackled by means of genetic programming \citep{koza1992genetic, Schmidt2009,Staelens2013,Arnaldo2015,Bladek2019}.
Unfortunately, they have to be carefully designed for each problem. Also, predicting formulas was slow and computationally expensive. For a single problem, an entire population of formulas had to be evolved and evaluated repeatedly through many generations.
In recent years neural approaches emerged \citep{peterson19deep, mundhenk21symbolic} to tackle these problems.
They were trained end-to-end with the sampled points as the input and the symbolic representation of the formula as the output.
However, these methods are still trained from scratch for each formula. Although they work well for simple formulas,
they are impractically slow and inefficient for more complex formulas due to their reliance on reinforcement learning.
To increase the efficiency, fully supervised approaches were proposed \citep{biggio21neural,  valipour21symbolicgpt}, that train a transformer model on a large collection of formulas. 
The formula is autoregressively generated by predicting each symbol conditioned on previously decoded symbols.
The generated expression is decoded without constants, i.e., all constants were replaced by a special symbol, and the concrete values
were found using global optimization. 
We argue that the concrete values of constants have a large impact on the generated function, and without
predicting them, the model will not learn to represent the data well. For example, we can imagine a simple
model returning a sum of sine and cosine functions with increasingly higher frequencies. Since every function can be
expressed using the Fourier basis, by changing the constants, we are able to represent most functions with low error.

Inspired by \citet{dascoli22deep}, where a similar idea was applied to the problem of recurrent sequences,
we propose a novel approach called the \ours that generates the concrete values for constants alongside
the symbolic representation of the formula. More specifically, we introduce the following contributions:
\begin{itemize}
\item We design a transformer-based architecture trained end-to-end on a large set of formulas consisting of hundreds of millions of formulas.
\item The model generates both the symbolic representation of the formula and the concrete values for
    all constants at the same time. This allows the symbolic decoder to condition on the generated constants
    and it improves the quality of the symbolic representation.
\item We also use the generated constants to initialize the local gradient search to fine-tune the final
    constants effectively and reliably.
\item Our approach was thoroughly evaluated and compared to relevant methods. Also, we validate our design choices
    in an ablation study.
\item The source code and the pre-trained model checkpoints are publicly available \footnote{\url{https://github.com/vastlik/symformer}}.
\end{itemize}

\section{Related work}

\textbf{Genetic Programming approaches} are a traditional way of solving SR \citep{koza1992genetic}. Genetic programming evolves expressions encoded as trees using selection, crossover, and mutation. A limitation of the genetic algorithm-based approaches is that they are sensitive to the choice of hyperparameters \citep{peterson19deep}. They need to evolve each equation from scratch, which is slow, and the models tend to increase in complexity without much performance improvement. It is also problematic to tune expression constants only by using genetic operators. 

\textbf{Neural Network approaches} can be generally divided into three categories. The first one is approaches based on Equation learner (EQL) \citep{martius16extrapolation, sahoo18learning, werner21informed}. The idea behind EQL is to find function $f(x)=y$ by training a neural network on $x$ as input and $y$ values as output while using as few network weights as possible. As activation functions, elementary functions ($\sin$, $\log$, \ldots) are used, and after the training, they are read from the network with corresponding weights. A limitation of such an approach is that they require special handling of functions that are not defined over the whole $\mathbb{R}$ (e.g., $\log$), that the depth of the network limits the complexity of the predicted equation. Lastly, they can be slow since they need to find each equation from scratch.

The second approach is based on training a recurrent neural network (RNN) using reinforcement learning \citep{peterson19deep}. The idea is to let the RNN generate the equation and then calculate the reward function as an error between the ground truth $f(x)$ values and the values from the predicted function $\hat{f}(x)$. An interesting extension is proposed by \citet{mundhenk21symbolic}, where they sample from the RNN, and the output is then taken as an initial population for a genetic algorithm. The limitations of both of these approaches are that the model does not predict the constants, and therefore they have to be found through global optimization in postprocessing steps which slows down the whole training loop \citep{mundhenk21symbolic, peterson19deep}.

The transformer-based approach is proposed by \citet{valipour21symbolicgpt, biggio21neural, dascoli22deep}, where they first generate a large amount of training data and train a transformer \citep{vaswani17attention} model in a supervised manner. \citet{valipour21symbolicgpt} train a GPT-2 \citep{radford19language} model on pairs of points and symbolic output. Then they use global optimization to find the constants for each equation. \citet{biggio21neural} uses the encoder from the Set transformer \citet{lee18set} and the decoder from original transformer architecture \citep{vaswani17attention}. Similar to \citet{valipour21symbolicgpt}, they train the models only on skeletons (expression  without the constants), and afterward, they fit the constants using global optimization. Another extension is introduced by \citet{dascoli22deep}, where they train the transformer model \citep{vaswani17attention} on recurrent sequences. They predict the expression constants jointly by encoding them into the symbolic output. To encode integers, they use their base $b$ representation e.g., for $x = -325$ and base $b = 30$ the representation would be $[-, 10, 25]$. In the case of floats, they use the IEEE 754 float representation and round the mantissa to the four most significant digits. They also introduce new tokens representing exponents. For example if we have number $-0.015$, then they encode it as $[-, 15, \mathrm{e}{-4}]$. The disadvantage of this approach is that the mantissa has only finite precision. Therefore, the model typically only predicts the largest terms when approximating complicated functions \citep{dascoli22deep}.

\section{Method}
\begin{figure}[t!]
    \centering
    \includegraphics[width=\textwidth]{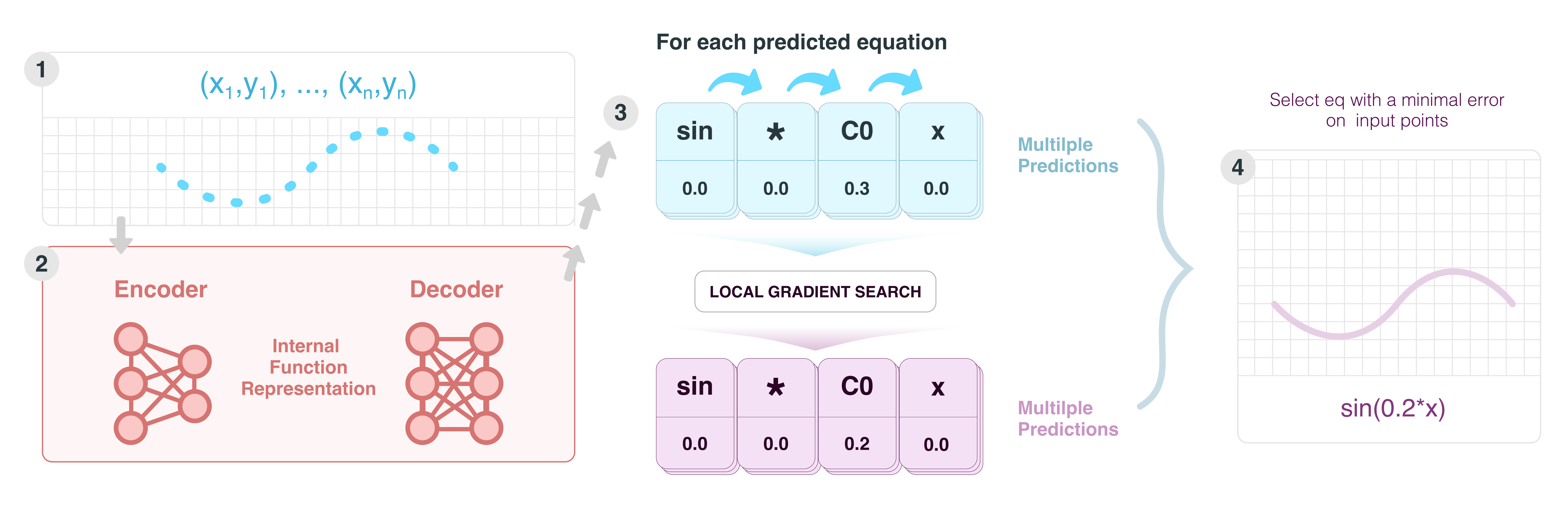}
    \caption{Schematic diagram of inference. The input points are passed through the transformer, generating several candidate equations using Top-K sampling. These candidates are further improved using gradient descent. The final equation is then selected by the lowest mean squared error. }
    \label{fig:model_architecture}
\end{figure}

In symbolic regression, it is assumed that there is an unknown function $f$ and that we observe its output on a finite set of input points.
The goal is then to find the mathematical formula of this function. Therefore, we want to find a function $\hat{f}$ such that the squared difference between the function's output on the input points and the outputs of the unknown function $f$ is minimized.

Given a set of observed input-output pairs, our model generates the symbolic representation of the formula together with values of all constants present in the formula in a single forward pass of a neural network.
This is visualized in \Cref{fig:model_architecture}.
First, we transform all input-output pairs using an encoder block to obtain an internal representation.
Given this representation, the decoder then autoregressively generates individual symbols and corresponding constants. 
This means that in order to generate the next token in the symbolic representation of the function, we pass all previously generated tokens and constants to the decoder.
We proceed in this fashion until we obtain the entire formula.
During inference, we sample multiple predictions from the model and fine-tune all constants to minimize the error between the predicted formula's outputs and the observed outputs.
Finally, we select the prediction with the lowest error.

\subsection{Model architecture}

Our neural network is an autoregressive transformer-based architecture that contains an encoder with cross-attention blocks \citep{lee18set,jaegle21perceiver} and a simpler self-attention decoder \citep{vaswani17attention}.
The input to the encoder consists of the data points, which are first passed through a trainable affine layer to project them into a latent space.
The resulting vectors are then passed through several induced set attention blocks \citep{lee18set}, which are two cross attention layers.
First, cross attention uses a set of trainable vectors as the queries and the input features as keys and values.
Its output is used as the keys and values for the second cross attention, and the original input vectors are used as the queries.
After these cross attention layers, we add a dropout layer \citep{srivastava14dropout}.
In the end, we compute cross attention between a set of trainable vectors (queries) to fix the size of the output such that it does not depend on the number of input points. This final representation is then passed to the decoder.

The decoder autoregressively generates the symbolic representation and the constants given the encoder's representation.
The input symbols are first passed through the embedding layer and then pairwise summed with trainable positional encoding vectors.
These newly formed vectors are pairwise concatenated with affine-projected constants. The resulting vectors are then passed through several decoder layers \citep{vaswani17attention}.
The decoder has two heads, where the first one is a classification head which predicts the probability distribution over the next symbol in the sequence.
If the predicted symbol is a constant, the other (regression) head outputs its value.
Training can efficiently process each sequence in a single forward pass of the network thanks to the masked attention and teacher forcing \citep{vaswani17attention}.

\subsection{Training \& inference}
We train the model using cross-entropy $\mathcal{L}_{class}$ for the symbolic expression and mean squared error $\mathcal{L}_{\text{MSE}}$ for the constants:
\begin{equation}
   \mathcal{L}=\mathcal{L}_{class} + \lambda \mathcal{L}_{\text{MSE}},
\end{equation}
where $\lambda$ is a hyperparameter, If $\lambda$ is too small, the model will not learn to predict the constants at all, and if $\lambda$ is too large, the model will not learn to predict the symbolic output well, and therefore, the constants will be useless. 
Therefore, at the beginning of the training, we set $\lambda$ to zero, and after a few epochs, we gradually increase it using the cosine schedule \citep{loshchilov16sgdr}. Note that we calculate the regression loss only at indices where the model should predict a constant. We have also found it beneficial to add a small random noise sampled from $\mathcal{N}(0, \sigma^2)$ to the constants during the training since, during inference, the constants are not always precise. Parameter $\sigma$ is decreased according to the cosine schedule.

During inference, we use Top-K sampling \citep{fan18hierarchical} to generate candidate equations. Then, we fix the symbolic expression and run gradient descent on all constants. We use the mean squared error between the predicted function's outputs and the outputs of the ground-truth function. Finally, we select the equation with the lowest error on the input points. 

\subsection{Dataset generation}
\label{sec:dataset_generation}
We generate two training datasets, one with 130 million equations containing only univariate functions and the second one with 100 million functions containing bivariate functions, by following the same algorithm as described by \citet{lample19deep} with the maximum of ten operators. The algorithm starts by generating a random unary-binary tree and filling the nodes with appropriate operations. The unnormalized probabilities of each operation and operator and the hyperparameters of the generator are given in \supp In our dataset, we have also introduced new operators such as $\text{pow2} (\cdot)=(\cdot)^2$, $\text{pow3} (\cdot)=(\cdot)^3$, \ldots to make it easier to represent them. The generated expressions are then simplified using SymPy \citep{meurer17sympy}. We discard the expressions that cannot be simplified in 5 seconds. Finally, we sample uniformly at random 100 points (200 for bivariate functions) from the interval $[-5,5]$, and if there are any non-finite values (NaN or $\pm \infty$), we try $(0,5]$ and then $[-5,0)$ (similarly for the bivariate dataset). The reason for selecting these intervals is that functions such as $\log{x}$, $\sqrt{x}$ or $\log{-x}$ are not defined on the full interval. Furthermore, we ignore any equations with values on the sampled points larger than $10^7$ (in absolute value). Similarly, we discard equations with constants smaller than $10^{-10}$ or larger than $10^{10}$ in absolute value. We also ignore any linear functions created by the simplification process.
We do not want to keep all the linear functions because the dataset would be biased towards linear functions.
Finally, we throw away any constant functions and functions that contain more than 50 symbolic tokens.

\subsection{Expression encoding}
\label{sec:expression_encoding}

We use preorder tree representation to encode expressions and replace constants with special symbols. The constants are encoded using a scientific-like notation where a constant $C$ is represented as a tuple of the exponent $c_e$ and the mantissa $c_m$:
\begin{equation}
C \approx c_{m} \cdot 10^{c_{e}}, \quad\quad c_e = \lceil \log_{10}{C} \rceil, \quad\quad c_m = \frac{C}{10^{c_e}}\,.
\end{equation}
In this representation, the mantissa is in the range $[-1, 1]$, and the exponent is an integer.
For example the expression $0.017 \cdot x + 1781.5$ will have symbols [+, mul, x, C-1, C4] and constants [0, 0, 0, 0.17, 0.17815].
To further help the model represent constants, we add all integers from interval $[-5,5]$ into the model vocabulary. 
Different encodings are compared in Section \ref{sec:ablation_study}. In contrast to \citet{dascoli22deep}, who are able to express constants only up to four most significant digits, our approach achieves full float precision.

\section{Experiments}
\label{sec:experiments}
This section describes our training setting and the metrics that we used to demonstrate the model's ability to predict the formulas and compare our model to previous approaches. We also show how the \ours generalizes to two dimensions and outside of the known range. Furthermore, we manually inspect the model's predictions to examine different equivalent mathematical formulas that the \ours found. In the end, we compare different encodings and their impact on the model's performance. In our experiments, we refer to the model trained only on univariate functions as the Univariate \ours and the model trained on both the univariate and bivariate functions as the Bivariate \ours. We also always use a local gradient search on the constants if not stated otherwise.

\subsection{Training}
We train our model using the Adam optimizer \citep{kingma14adam} for 3 epochs on 8 NVIDIA A100 GPUs. The training of the model takes roughly 33 hours. We use 130 million univariate equations for the training set and 10\,000 for the validation set. Furthermore, we randomly selected 256 equations to calculate the metrics using the beam search. We use a training schedule similar to the original transformer \citep{vaswani17attention}. However, we divide the learning rate by five since the training often diverged when using the original learning rate. The regression $\lambda$ is set according to the cosine schedule and delayed for 97\,700 gradient steps, reaching 1.0 at the end of the training.\footnote{The regression $\lambda$ and the learning rate were updated every $10^6/1024\approx977$ gradient steps (batch size was 1\,024).}
For the random noise, we sample from $\mathcal{N}(0, \epsilon)$, where $\epsilon$ is initially set to $0.1$ and decreased to zero during training using the same schedule. The complete set of hyperparameters for the model containing approximately 95 million parameters can be seen in \supp The hyperparameters were found empirically. We use the same settings for our Bivariate \ours.

\subsection{Metrics}
To assess the quality of the model, we have selected two metrics: the relative error and the coefficient of determination ($R^2$).
The relative error is the average absolute difference between the predicted value and the ground truth divided by the absolute value of the ground truth:
\begin{equation}
    \text{RE}(y,\hat{y})=\frac{1}{k}\sum_{i=1}^{k}\bigg|\frac{y_i-\hat{y_i}}{y_i}\bigg|,
\end{equation}
where $y_i$ and $\hat{y_i}$ are the ground-truth and predicted values for point $i$, respectively.
The coefficient of determination ($R^2$) \citep{glantz00primer} is defined as follows:
\begin{equation}
    R^2(y,\hat{y})=1-\frac{\sum_{i=1}^{k} (y_i - \hat{y}_i)^2}{\sum_{i=1}^{k} (y_i - \bar{y})^2},
\end{equation}
where $y_i$ and $\hat{y_i}$ are the ground-truth and predicted values for point $i$, respectively. $\bar{y}$ is the average of $y_i$ over all the points. The advantage of using $R^2$ is its nice interpretation. If $R^2>0$, then our prediction is better than the predicting just the average value and if $R^2=1$, then we have a perfect model.

\begin{figure}[htbp]
\definecolor{figblue}{HTML}{1f77b4}
\definecolor{figorange}{HTML}{d96e0f}
    \captionsetup[subfigure]{aboveskip=-1pt,belowskip=-1pt,font=footnotesize,justification=centering}
     \centering
     \begin{subfigure}[b]{0.48\textwidth}
         \centering
         \caption{\textcolor{figblue}{GT}: $(1+x^{-2})^{-0.5}$, \textcolor{figorange}{Pred}: $\sin(|\mathrm{atan}(x)|)$}
         \includegraphics[width=\textwidth]{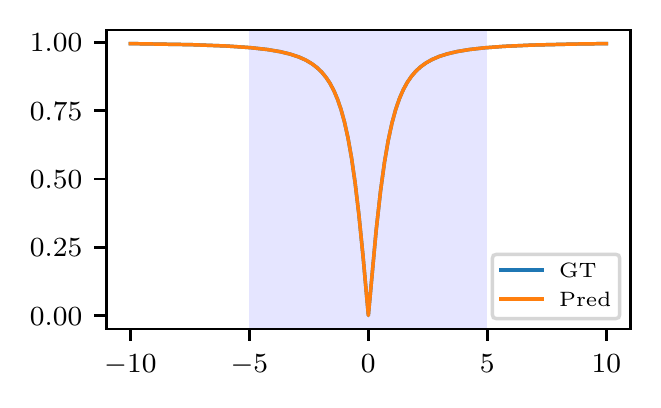}
     \end{subfigure}
     \begin{subfigure}[b]{0.48\textwidth}
         \centering
         \caption{\textcolor{figblue}{GT}: $7.7 + 3 \ln(x)$, \textcolor{figorange}{Pred}: $3 \ln(x) + 7.9$}
         \includegraphics[width=\textwidth]{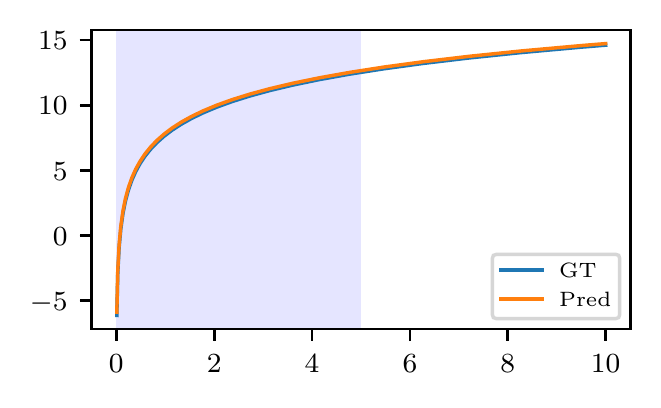}
    \end{subfigure}
    \begin{subfigure}[b]{0.48\textwidth}
         \centering
         \caption{\parbox{0.68\textwidth}{\textcolor{figblue}{GT}: $-7.46-0.8x+x\cos(\tan(x))$,\\ \textcolor{figorange}{Pred}: $-7.7-x+x\cos(\tan(x))$}}
         \includegraphics[width=\textwidth]{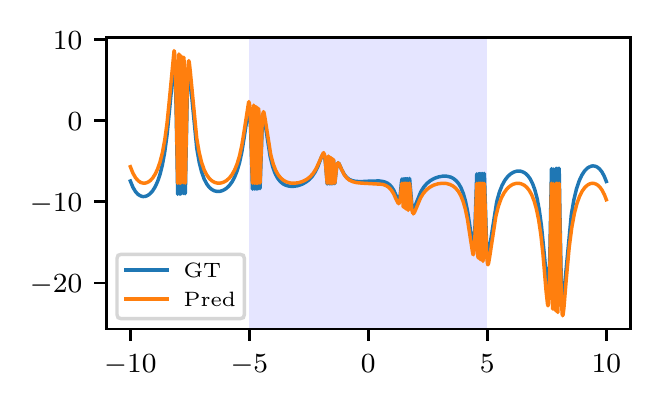}
    \end{subfigure}
    \begin{subfigure}[b]{0.48\textwidth}
         \centering
         \caption{\textcolor{figblue}{GT}: $0.2\cos(4x)$, \textcolor{figorange}{Pred}: $\mathrm{asin}(0.2\cos(4x))$}
         \includegraphics[width=\textwidth]{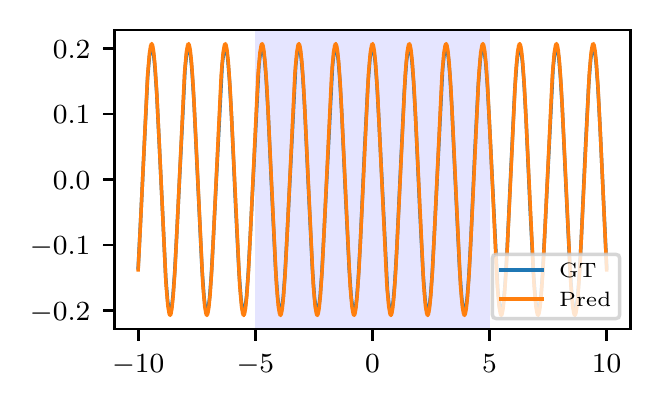}
    \end{subfigure}
    \begin{subfigure}[b]{0.48\textwidth}
         \centering
         \caption{\parbox{0.68\textwidth}{\textcolor{figblue}{GT}: $-60.9 \cdot x \cdot \exp(-x)$,\\ \textcolor{figorange}{Pred}: $0.002x^3 - 61.2 \cdot x \cdot \exp(-x)$}}
         \includegraphics[width=\textwidth]{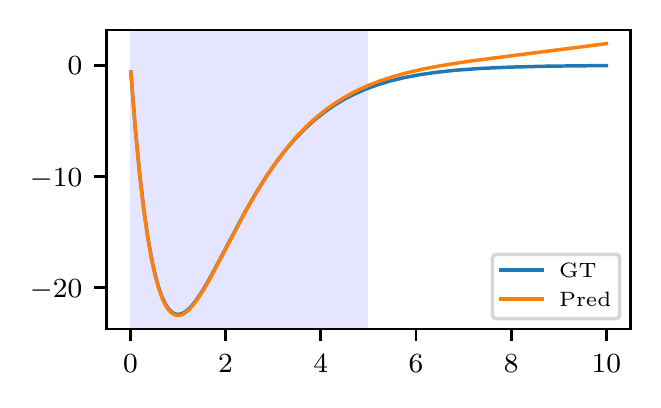}
    \end{subfigure}
    \begin{subfigure}[b]{0.48\textwidth}
         \centering
         \caption{\parbox{0.75\textwidth}{\textcolor{figblue}{GT}: $0.34x+(((x)^2)+\sin{(0.96+x)}$,\\ \textcolor{figorange}{Pred}: $x^2 + \cos(x) + \mathrm{atan}(x)$}}
         \includegraphics[width=\textwidth]{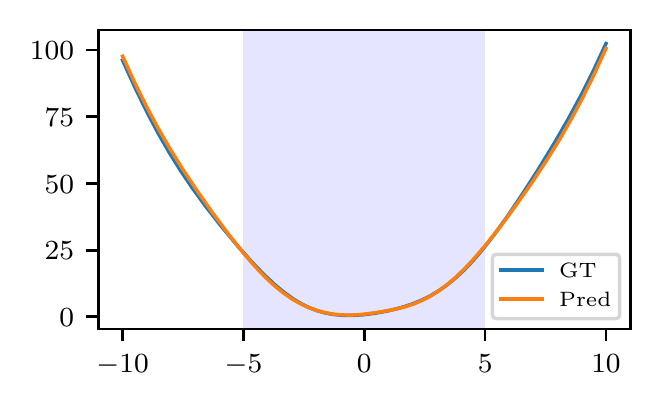}
    \end{subfigure}
        \begin{subfigure}[b]{\textwidth}
         \centering
         \caption{\textcolor{figblue}{GT}: $x-x^3+y^{-1}\sin{(y)}$,\quad\quad\quad\quad\quad\quad \textcolor{figorange}{Pred}: $x-x^3+y^{-1}\sin{(y)}$\label{fig:model_predictions:3d-plot}}
         \includegraphics[width=0.38\textwidth, clip, trim=0 0 0 15]{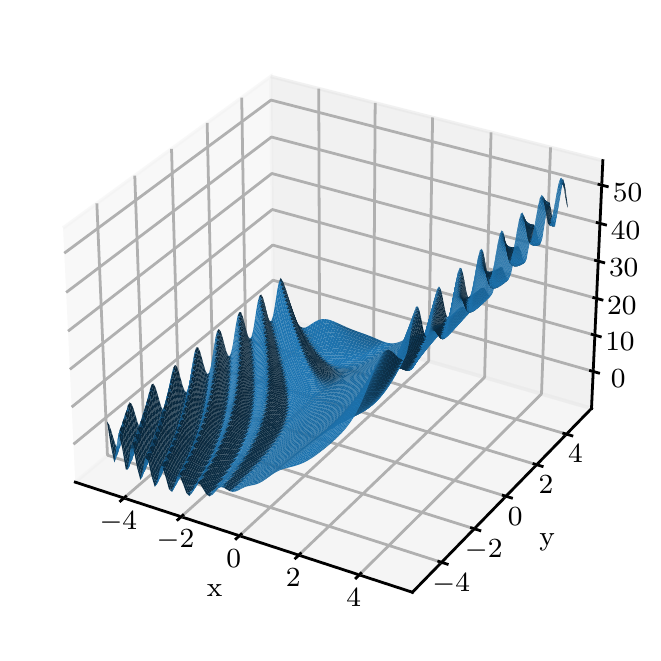}
         \includegraphics[width=0.38\textwidth, clip, trim=0 0 0 15]{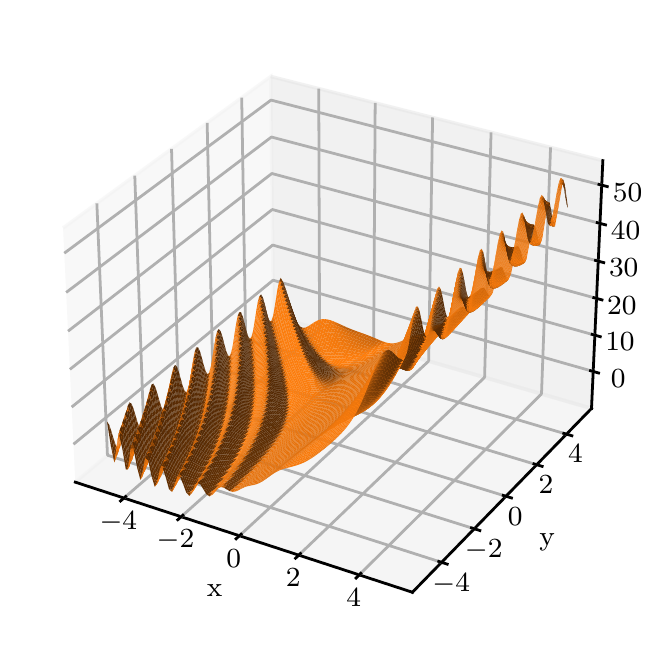}
    \end{subfigure}
    \caption{Examples of model predictions using Top-K sampling with $K=20$ and 256 samples. The shaded area represents the sampling range. For the 3D functions, $x$ and $y$ were sampled from $[-5,5]$. `\textcolor{figblue}{GT}' denotes ground-truth and `\textcolor{figorange}{Pred}', the model prediction. The first six images are generated using univariate SymFormer, and the last one was generated by bivariate SymFormer.}
        \label{fig:model_predictions}
\end{figure}

\subsection{In-domain performance}
To demonstrate the \ours's ability to predict the formulas successfully, we use Top-K sampling \citep{fan18hierarchical} with $K=16$ and 256 samples to generate the best equation. In our experiments, we report the median values of all metrics since the mean can be skewed by outliers. In Figure~\ref{fig:model_predictions}, we plot some of the model predictions.
The Univariate \ours achieved an $R^2$ of 0.9995 and a relative error of 0.0288.
Furthermore, when we used the local gradient search, the model improved to $R^2$ 1.0000 and a relative error of 0.0010. 
The Bivariate \ours achieved an $R^2$ of 0.9996 and a relative error of 0.0389 using Top-K \citep{fan18hierarchical} with $K=16$ and 1.0000 $R^2$ and a relative error of 0.0035 when the local gradient search was used. This demonstrates the model's ability to generalize to higher dimensions.

\subsection{Comparison to previous approaches}
\label{sec:benchmark}

To compare our results to current state-of-the-art approaches, we use the Nguyen benchmark \citep{nguyen11semantically}, R rationals \citep{krawiec13approximating}, Livermore, \citep{peterson19deep}, Keijzer \citep{keijzer03improving}, Constant, and Koza \citep{koza94genetic}. The complete benchmark functions are given in \supp
Unfortunately, these benchmarks are better suited for methods where parameters such as the number of variables, set of symbols, or sampling range are set specifically to match the problem at hand.
In our method, these parameters are fixed in the beginning and cannot be changed later.
Note that it is difficult to make a completely fair comparison on the benchmarks for two reasons. The first one is that some methods use a restricted vocabulary and thus have a smaller search space giving them an advantage over our method. The second problem arises from the different sampling ranges and the number of sampled points.

\begin{table}[t]
    \caption{Results comparing the \textbf{\ours} with state-of-the-art methods on several benchmarks. The \textbf{\Ours} uses Top-K sampling with $K=20$ while generating 1024 samples and improving them using local gradient search with early stopping. We report $R^2$ and the average time to generate an equation in seconds.
    \label{tab:benchmark_results}\\}
    \centering
    \begin{tabular}{
@{}
l@{\extracolsep{0.1cm}}
cD{.}{.}{3.4}@{\extracolsep{0.1cm}}
cD{.}{.}{3.4}@{\extracolsep{0.1cm}}
cD{.}{.}{3.4}@{\extracolsep{0.1cm}}
@{}}

\setlength\tabcolsep{0.05cm}

& \multicolumn{2}{c}{\textbf{SymFormer}} & \multicolumn{2}{c}{NSRS \citep{biggio21neural}} & \multicolumn{2}{c}{DSO \citep{mundhenk21symbolic}} \\
\cmidrule{2-3} \cmidrule{4-5} \cmidrule{6-7}
Benchmark & \multicolumn{1}{c}{$R^2\uparrow$} & \multicolumn{1}{c}{Time\,(s)\,$\downarrow$} & \multicolumn{1}{c}{$R^2\uparrow$} & \multicolumn{1}{c}{Time\,(s)\,$\downarrow$} & \multicolumn{1}{c}{$R^2\uparrow$} & \multicolumn{1}{c}{Time\,(s)\,$\downarrow$} \\
\cmidrule{1-1} \cmidrule{2-3} \cmidrule{4-5} \cmidrule{6-7}
 Nguyen       & \textbf{0.99998}    & 47.50            & 0.96744 & 169.46 & 0.99297 & 140.25 \\
 R & 0.99986  & 94.33               & \textbf{1.00000} & 95.67 & 0.97488 & 855.33 \\
 Livermore    & \textbf{0.99996}    & 43.00            & 0.88551 & 193.09 & 0.99651 & 276.32 \\
 Koza         & \textbf{1.00000}          & 101.00     & 0.99999 & 111.50 & \textbf{1.00000} & 217.50 \\
 Keijzer      & \textbf{0.99904}    & 48.67            & 0.97392 & 255.50 & 0.95302 & 3929.50 \\
 Constant     & 0.99998             & 90.88            & 0.88742 & 230.38 & \textbf{1.00000} & 2816.19 \\
\cmidrule{1-1} \cmidrule{2-7}
 Overall avg. & \textbf{0.99978} & 52.95 & 0.92901 & 199.63 & 0.99443 & 326.53  \\
 \bottomrule
\end{tabular}

\end{table}

We use Top-K sampling with $K=20$ and 1\,024 samples with early stopping for the benchmark. From the results in Table~\ref{tab:benchmark_results} we can see that the \ours method is competitive in terms of the model performance on all of the benchmarks while outperforming both NSRS \citep{biggio21neural} and DSO \citep{mundhenk21symbolic} in the time required to find the equation. One of the observations that we have found is that the model sometimes predicts semantically the same expression as the ground-truth, but using a more complex expression, \eg, in one case model had to predict $\frac{x^2+x}{2}$, but predicted $\ln{(\exp(0.5x^2)\exp{(0.5x)})}$, which is same after simplification, but unreasonably complex.
This was likely caused by the distribution of our dataset.

\begin{table}[t]
    \caption{Results comparing the \textbf{\ours}, when the different model is used for univariate functions and bivariate functions and when the Bivariate \ours is used for all the benchmark functions. The (Bivariate) \textbf{\Ours} uses Top-K sampling with $K=20$ while generating 1024 samples and improving them using local gradient search with early stopping. We report $R^2$ and the average time to generate an equation in seconds.
    \label{tab:benchmark_results_1d_2d}\\}
    \centering
    \begin{tabular}{
@{}
l@{\extracolsep{0.1cm}}
cD{.}{.}{3.4}@{\extracolsep{0.1cm}}
cD{.}{.}{3.4}@{\extracolsep{0.1cm}}
@{}}

\setlength\tabcolsep{0.05cm}

& \multicolumn{2}{c}{SymFormer} & \multicolumn{2}{c}{Bivariate SymFormer} \\
\cmidrule{2-3} \cmidrule{4-5}
Benchmark & \multicolumn{1}{c}{$R^2\uparrow$} & \multicolumn{1}{c}{Time\,(s)\,$\downarrow$} & \multicolumn{1}{c}{$R^2\uparrow$} & \multicolumn{1}{c}{Time\,(s)\,$\downarrow$}  \\
\cmidrule{1-1} \cmidrule{2-3} \cmidrule{4-5}
 Nguyen       & \textbf{0.99998}    & 47.50     & 0.99996 & 139.46  \\
 R            & \textbf{0.99986}    & 94.33     & 0.99985 & 418.67 \\
 Livermore    & \textbf{0.99996}    & 43.00     & 0.99992 & 170.00 \\
 Koza         & \textbf{1.00000}    & 101.00    & \textbf{1.00000} & 81.50 \\
 Keijzer      & \textbf{0.99904}    & 48.67     & 0.99884 & 250.66 \\
 Constant     & \textbf{0.99998}             & 90.88     & 0.99997 & 188.50 \\
\cmidrule{1-1} \cmidrule{2-5}
 Overall avg. & \textbf{0.99978} & 52.95 & 0.99946 & 174.32 \\
 \bottomrule
\end{tabular}

\end{table}

Furthermore, to demonstrate the Bivariate \ours performance on both the univariate and bivariate functions, we evaluated a single model on all univariate and bivariate functions using the same benchmark.
Note that the benchmark functions are mostly univariate. In Table \ref{tab:benchmark_results_1d_2d}, we can notice only a slight drop in performance. However, the average inference time increased, which could be explained by a larger search space the model needed to handle during the optimisation of constants. Furthermore, we have manually inspected the model's predictions on benchmark functions. We found that the model had no problems recovering simple equations but was slow or failed in cases of more complicated functions.

\subsection{Out-of-domain performance}

\begin{figure}
     \centering
    \includegraphics[width=0.48\textwidth]{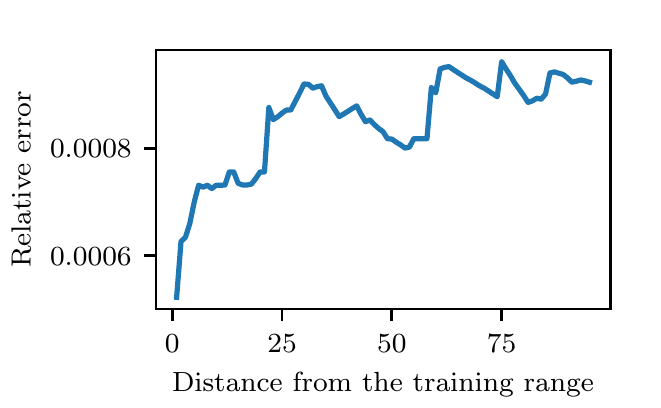}
    \includegraphics[width=0.48\textwidth]{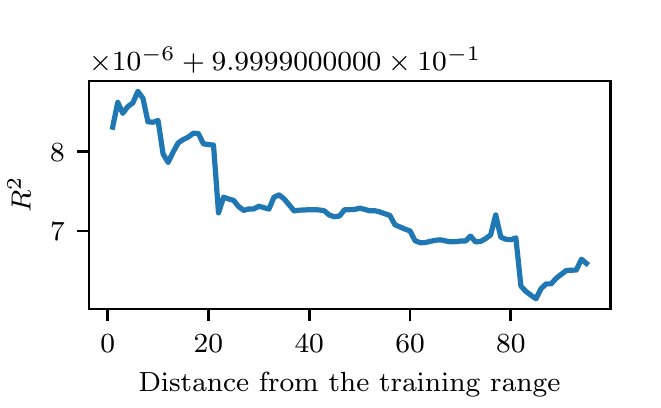}
    \caption{The effect of a distance when calculating the relative error outside the sampling range. They are estimated on 1024 equations generated with Top-K sampling with $K=20$ and 256 samples. Each of the equations is also improved by a local gradient search on constants.}
    \label{fig:grad_re}
\end{figure}
One intriguing property of symbolic regression is its ability to predict the correct values outside the sampling range.
To test it, we first run the inference on the points sampled from the training range and then evaluate these predicted functions on points outside the sampling range.
More formally we calculate the metrics on the function values for points sampled from the set $\{x \in \mathbb{R} \vert 5 < |x| < 5 + d\}$, where $d$ is the maximal distance.
The effect of the distance on the relative error and $R^2$ can be seen in \Cref{fig:grad_re}.
Even though the error increases with the distance, the final relative error is minimal, even for the maximal range.
Therefore, we can conclude that the model generalises outside of the sampling range, and the local gradient search does not overfit the constants to the sampled data.

\subsection{Discovering mathematically equivalent expressions}
To qualitatively evaluate the \ours's prediction capabilities, we have manually inspected the model's predictions.
The model is often able to find mathematically equivalent expressions.
In one case the model discovered the rule $-\ln{x}=\ln{x^{-1}}$. The goal was to predict $-\frac{8}{\ln{x}}$, but the model predicted $\frac{8}{\ln{(x^{-1})}}$. Another rule the model discovered was the law of exponents $(x^a)^b=x^{a \cdot b}$. It was observed when the model had to predict $(x^{-1})^{1.5}$, but found an equivalent form $\frac{1}{\sqrt{x^3}}$.
Furthermore, the model was also able to find some trigonometric equivalencies such as $\frac{1}{\tan{x}}=\cot{x}$.
However, a more interesting example is the expression $\cos{(3.5 + 2x + x^2)}$.
In this case, the model predicted $-\sin{(2+2x+x^2)}$, which has a very small error.
The rule, that the model discovered is $\cos{(\frac{\pi}{2}+x)}=-\sin{(x)}$.
Applying this rule we get $ \cos(3.5+2x+x^2) \approx \cos(\frac{\pi}{2} + 1.9 + 2x + x^2) \approx -\sin(1.9 + 2x + x^2)$.
Another interesting example the model predicted is the exponential rule $a^x=e^{x\cdot \ln{a}}$.
This rule is probably used when the \ours needs to deal with precise constants, due to the numerical stability of $\ln$.
For example, the model had to find $(-1.3673x)$, but it found $\ln(0.2492^x)=\ln(e^{x\ln{0.2492}})=x\ln{0.2492}=-1.3895x$ which is close to the previous expression. 

\subsection{Ablation study}
\label{sec:ablation_study}
\begin{table}[t!]
    \centering
        \caption{Comparison of expression encoding strategies and local gradient search (`LGS'). The \textbf{\Ours} uses both `extended encoding' and local gradient search. Metrics are estimated using 256 equations using Top-K sampling with $K=20$ and 256 samples, where the equation with the lowest mean squared error on the input points is selected. We report the $R^2$ and the relative error. The base encoding refers to case when no preprocessing for the constants is used. BFGS init refers to a situation when the predicted constants are used as a starting point for the BFGS \citep{fletcher87practical}. The 'GS' refers to the case when the gradient search was used to find or improve the constants further. \label{tab:ablation_study}\\}
\vspace{0.1cm}
\begin{tabular}{
@{}
l@{}
c@{\extracolsep{0.1cm}}
c@{\extracolsep{0.1cm}}
c@{\extracolsep{0.1cm}}
@{}
}
\toprule
Model & GS & $R^2$ $\uparrow$  & Relative error $\downarrow$ \\
\cmidrule{1-2} \cmidrule{3-3} \cmidrule{4-4}
No constants + BFGS  &  \cmark            & 0.9929          & 0.1547          \\
Base encoding    &  \xmark              & 0.9979          &  0.0669         \\
Base encoding + LGS    &  \cmark             &  \textbf{1.0000} & 0.0089          \\
Extended encoding & \xmark              & 0.9995          & 0.0288          \\
Extended encoding  + BFGS init & \cmark               & 0.9998          & 0.0303         \\
\cmidrule{1-2} \cmidrule{3-4}
\textbf{\Ours} & \cmark  & \textbf{1.0000} & \textbf{0.0010} \\
\bottomrule
\end{tabular}

\end{table}
This ablation study aims to look at the effect of different constant encodings. In the first setting, we did not predict the constants and used a global optimisation (BFGS \citep{fletcher87practical}) to find the constants.
This setting is the same as used by \citet{biggio21neural}.
In the second setting we used the constants during training, but we did not preprocess them.
Therefore, a single symbol, `const', was used to represent any constant regardless of its magnitude.
In the last case, we have used encoding as described in \Cref{sec:expression_encoding},
while also trying to use the predicted values of constants as a starting point for global optimisation.
From the results in Table \ref{tab:ablation_study}, we can see that the constants help the model performance in terms of both the $R^2$ and the relative error. Therefore, one can conclude that the performance of SymFormer in comparison to \citet{biggio21neural} is better not because of a different dataset or a larger model but because of the usage of constants during training.
The last row shows the results for the extended encoding, which uses a local gradient search to improve the constants further. 
The extended encoding clearly outperforms the base encoding in terms of both the $R^2$ and the relative error.
We believe this to be the case because it is easier for the model to attend to previous symbolic tokens than to real values and, therefore, the model can make a more informed decision when predicting the next symbol in the sequence.

\section{Conclusion}
\label{sec:conclusion}
To tackle the problem of symbolic regression, we introduced a novel transformer-based approach called the \ours
that uses a neural network trained on hundreds of millions of formulas to be able to generate 
a symbolic representation of a previously unseen formula given a set of input-output pairs efficiently.
Our model jointly predicts the symbolic representation of a function and the values of all constants in a single forward pass of a neural network.
A local gradient search is used to improve constants further to fit the input points better. 
We demonstrated that the \ours is competitive with current state-of-the-art approaches while outperforming them in terms of the time required to find the expression.
We validated the importance of the proposed encoding of constants in an ablation study.
Furthermore, by evaluating the \ours outside the sampling range, we showed that it has good extrapolation capabilities.
Finally, in a qualitative evaluation, we present some intriguing mathematical relations the model was able to recover just by learning on a large collection of formulas.

\textbf{Limitations.} One of the limitations of our approach is that the maximal number of dimensions and the sampling range cannot be changed after the model was trained. In the future, this can be partially tackled by varying the sampling distribution of the input points during training.

\begin{ack}
This work was supported by the European Regional Development Fund under the project Robotics for Industry 4.0 (reg. no. CZ.02.1.01/0.0/0.0/15\_003/0000470). Jonáš Kulhánek was funded under project IMPACT (reg. no. CZ.02.1.01/0.0/0.0/15\_003/0000468) and by the Grant Agency of the Czech Technical University in Prague (grant no. SGS22/112/OHK3/2T/13). Computational resources were further funded by the Ministry of Education, Youth and Sports of the Czech Republic through the e-INFRA CZ (ID:90140). 
\end{ack}

{
\small
\bibliography{bibliography} 

\begin{thebibliography}{36}
\providecommand{\natexlab}[1]{#1}
\providecommand{\url}[1]{\texttt{#1}}
\expandafter\ifx\csname urlstyle\endcsname\relax
  \providecommand{\doi}[1]{doi: #1}\else
  \providecommand{\doi}{doi: \begingroup \urlstyle{rm}\Url}\fi

\bibitem[Arnaldo et~al.(2015)Arnaldo, O’Reilly, and
  Veeramachaneni]{Arnaldo2015}
Ignacio Arnaldo, Una-May O’Reilly, and Kalyan Veeramachaneni.
\newblock Building predictive models via feature synthesis.
\newblock In \emph{Proceedings of the 2015 Annual Conference on Genetic and
  Evolutionary Computation}, GECCO ’15, page 983–990, New York, NY, USA,
  2015. Association for Computing Machinery.
\newblock ISBN 9781450334723.
\newblock \doi{10.1145/2739480.2754693}.

\bibitem[Biggio et~al.(2021)Biggio, Bendinelli, Neitz, Lucchi, and
  Parascandolo]{biggio21neural}
Luca Biggio, Tommaso Bendinelli, Alexander Neitz, Aur{\'{e}}lien Lucchi, and
  Giambattista Parascandolo.
\newblock Neural symbolic regression that scales.
\newblock \emph{CoRR}, abs/2106.06427, 2021.
\newblock URL \url{https://arxiv.org/abs/2106.06427}.

\bibitem[B\l\k{a}dek and Krawiec(2019)]{Bladek2019}
Iwo B\l\k{a}dek and Krzysztof Krawiec.
\newblock Solving symbolic regression problems with formal constraints.
\newblock In \emph{Proceedings of the Genetic and Evolutionary Computation
  Conference}, GECCO '19, pages 977--984, New York, NY, USA, 2019. ACM.
\newblock ISBN 978-1-4503-6111-8.
\newblock \doi{10.1145/3321707.3321743}.

\bibitem[d'Ascoli et~al.(2022)d'Ascoli, Kamienny, Lample, and
  Charton]{dascoli22deep}
St{\'{e}}phane d'Ascoli, Pierre{-}Alexandre Kamienny, Guillaume Lample, and
  Fran{\c{c}}ois Charton.
\newblock Deep symbolic regression for recurrent sequences.
\newblock \emph{CoRR}, abs/2201.04600, 2022.
\newblock URL \url{https://arxiv.org/abs/2201.04600}.

\bibitem[Fan et~al.(2018)Fan, Lewis, and Dauphin]{fan18hierarchical}
Angela Fan, Mike Lewis, and Yann~N. Dauphin.
\newblock Hierarchical neural story generation.
\newblock \emph{CoRR}, abs/1805.04833, 2018.
\newblock URL \url{http://arxiv.org/abs/1805.04833}.

\bibitem[Fletcher(1987)]{fletcher87practical}
Roger Fletcher.
\newblock \emph{Practical Methods of Optimization}.
\newblock John Wiley \& Sons, New York, NY, USA, second edition, 1987.

\bibitem[Glantz and Slinker(2000)]{glantz00primer}
S.~Glantz and B.~Slinker.
\newblock \emph{Primer of Applied Regression \& Analysis of Variance}.
\newblock McGraw-Hill Education, 2000.
\newblock ISBN 9780071360869.
\newblock URL \url{https://books.google.cz/books?id=fzV2QgAACAAJ}.

\bibitem[Hein et~al.(2017)Hein, Udluft, and Runkler]{hein17interpretable}
Daniel Hein, Steffen Udluft, and Thomas~A. Runkler.
\newblock Interpretable policies for reinforcement learning by genetic
  programming.
\newblock \emph{CoRR}, abs/1712.04170, 2017.
\newblock URL \url{http://arxiv.org/abs/1712.04170}.

\bibitem[Jaegle et~al.(2021)Jaegle, Gimeno, Brock, Zisserman, Vinyals, and
  Carreira]{jaegle21perceiver}
Andrew Jaegle, Felix Gimeno, Andrew Brock, Andrew Zisserman, Oriol Vinyals, and
  Jo{\~{a}}o Carreira.
\newblock Perceiver: General perception with iterative attention.
\newblock \emph{CoRR}, abs/2103.03206, 2021.
\newblock URL \url{https://arxiv.org/abs/2103.03206}.

\bibitem[Keijzer(2003)]{keijzer03improving}
Maarten Keijzer.
\newblock Improving symbolic regression with interval arithmetic and linear
  scaling.
\newblock In Conor Ryan, Terence Soule, Maarten Keijzer, Edward Tsang, Riccardo
  Poli, and Ernesto Costa, editors, \emph{Genetic Programming}, pages 70--82,
  Berlin, Heidelberg, 2003. Springer Berlin Heidelberg.
\newblock ISBN 978-3-540-36599-0.

\bibitem[Kingma and Ba(2014)]{kingma14adam}
Diederik~P. Kingma and Jimmy Ba.
\newblock Adam: A method for stochastic optimization, 2014.
\newblock URL \url{http://arxiv.org/abs/1412.6980}.
\newblock cite arxiv:1412.6980Comment: Published as a conference paper at the
  3rd International Conference for Learning Representations, San Diego, 2015.

\bibitem[Koza(1992)]{koza1992genetic}
John~R. Koza.
\newblock \emph{Genetic Programming: {O}n the Programming of Computers by Means
  of Natural Selection}.
\newblock MIT Press, Cambridge, MA, USA, 1992.
\newblock ISBN 0-262-11170-5.

\bibitem[Koza(1994)]{koza94genetic}
John~R. Koza.
\newblock \emph{Genetic Programming II: Automatic Discovery of Reusable
  Programs}.
\newblock MIT Press, Cambridge, MA, USA, 1994.
\newblock ISBN 0262111896.

\bibitem[Krawiec and Pawlak(2013)]{krawiec13approximating}
Krzysztof Krawiec and Tomasz Pawlak.
\newblock Approximating geometric crossover by semantic backpropagation.
\newblock In \emph{Proceedings of the 15th Annual Conference on Genetic and
  Evolutionary Computation}, GECCO '13, page 941–948, New York, NY, USA,
  2013. Association for Computing Machinery.
\newblock ISBN 9781450319638.
\newblock \doi{10.1145/2463372.2463483}.
\newblock URL \url{https://doi.org/10.1145/2463372.2463483}.

\bibitem[Kubal{\'{\i}}k et~al.(2019)Kubal{\'{\i}}k, Zegklitz, Derner, and
  Babuska]{kubalik19symbolic}
Jir{\'{\i}} Kubal{\'{\i}}k, Jan Zegklitz, Erik Derner, and Robert Babuska.
\newblock Symbolic regression methods for reinforcement learning.
\newblock \emph{CoRR}, abs/1903.09688, 2019.
\newblock URL \url{http://arxiv.org/abs/1903.09688}.

\bibitem[Kubal{\'{\i}}k et~al.(2020)Kubal{\'{\i}}k, Derner, and
  Babuska]{kubalik20symbolic}
Jir{\'{\i}} Kubal{\'{\i}}k, Erik Derner, and Robert Babuska.
\newblock Symbolic regression driven by training data and prior knowledge.
\newblock \emph{CoRR}, abs/2004.11947, 2020.
\newblock URL \url{https://arxiv.org/abs/2004.11947}.

\bibitem[Lample and Charton(2019)]{lample19deep}
Guillaume Lample and Fran{\c{c}}ois Charton.
\newblock Deep learning for symbolic mathematics.
\newblock \emph{CoRR}, abs/1912.01412, 2019.
\newblock URL \url{http://arxiv.org/abs/1912.01412}.

\bibitem[Lee et~al.(2018)Lee, Lee, Kim, Kosiorek, Choi, and Teh]{lee18set}
Juho Lee, Yoonho Lee, Jungtaek Kim, Adam~R. Kosiorek, Seungjin Choi, and
  Yee~Whye Teh.
\newblock Set transformer.
\newblock \emph{CoRR}, abs/1810.00825, 2018.
\newblock URL \url{http://arxiv.org/abs/1810.00825}.

\bibitem[Loshchilov and Hutter(2016)]{loshchilov16sgdr}
Ilya Loshchilov and Frank Hutter.
\newblock {SGDR:} stochastic gradient descent with restarts.
\newblock \emph{CoRR}, abs/1608.03983, 2016.
\newblock URL \url{http://arxiv.org/abs/1608.03983}.

\bibitem[Martius and Lampert(2016)]{martius16extrapolation}
Georg Martius and Christoph~H. Lampert.
\newblock Extrapolation and learning equations.
\newblock \emph{CoRR}, abs/1610.02995, 2016.
\newblock URL \url{http://arxiv.org/abs/1610.02995}.

\bibitem[Matchev et~al.(2021)Matchev, Matcheva, and
  Roman]{konstantin21analytical}
Konstantin~T. Matchev, Katia Matcheva, and Alexander Roman.
\newblock Analytical modelling of exoplanet transit specroscopy with
  dimensional analysis and symbolic regression, 2021.
\newblock URL \url{https://arxiv.org/abs/2112.11600}.

\bibitem[Meurer et~al.(2017)Meurer, Smith, Paprocki, \v{C}ert\'{i}k, Kirpichev,
  Rocklin, Kumar, Ivanov, Moore, Singh, Rathnayake, Vig, Granger, Muller,
  Bonazzi, Gupta, Vats, Johansson, Pedregosa, Curry, Terrel, Rou\v{c}ka, Saboo,
  Fernando, Kulal, Cimrman, and Scopatz]{meurer17sympy}
Aaron Meurer, Christopher~P. Smith, Mateusz Paprocki, Ond\v{r}ej
  \v{C}ert\'{i}k, Sergey~B. Kirpichev, Matthew Rocklin, AMiT Kumar, Sergiu
  Ivanov, Jason~K. Moore, Sartaj Singh, Thilina Rathnayake, Sean Vig, Brian~E.
  Granger, Richard~P. Muller, Francesco Bonazzi, Harsh Gupta, Shivam Vats,
  Fredrik Johansson, Fabian Pedregosa, Matthew~J. Curry, Andy~R. Terrel,
  \v{S}t\v{e}p\'{a}n Rou\v{c}ka, Ashutosh Saboo, Isuru Fernando, Sumith Kulal,
  Robert Cimrman, and Anthony Scopatz.
\newblock Sympy: symbolic computing in python.
\newblock \emph{PeerJ Computer Science}, 3:\penalty0 e103, January 2017.
\newblock ISSN 2376-5992.
\newblock \doi{10.7717/peerj-cs.103}.
\newblock URL \url{https://doi.org/10.7717/peerj-cs.103}.

\bibitem[Mundhenk et~al.(2021)Mundhenk, Landajuela, Glatt, Santiago, Faissol,
  and Petersen]{mundhenk21symbolic}
T.~Nathan Mundhenk, Mikel Landajuela, Ruben Glatt, Cl{\'{a}}udio~P. Santiago,
  Daniel~M. Faissol, and Brenden~K. Petersen.
\newblock Symbolic regression via neural-guided genetic programming population
  seeding.
\newblock \emph{CoRR}, abs/2111.00053, 2021.
\newblock URL \url{https://arxiv.org/abs/2111.00053}.

\bibitem[Petersen(2019)]{peterson19deep}
Brenden~K. Petersen.
\newblock Deep symbolic regression: Recovering mathematical expressions from
  data via policy gradients.
\newblock \emph{CoRR}, abs/1912.04871, 2019.
\newblock URL \url{http://arxiv.org/abs/1912.04871}.

\bibitem[Radford et~al.(2019)Radford, Wu, Child, Luan, Amodei, and
  Sutskever]{radford19language}
Alec Radford, Jeff Wu, Rewon Child, David Luan, Dario Amodei, and Ilya
  Sutskever.
\newblock Language models are unsupervised multitask learners, 2019.

\bibitem[Sahoo et~al.(2018)Sahoo, Lampert, and Martius]{sahoo18learning}
Subham~S. Sahoo, Christoph~H. Lampert, and Georg Martius.
\newblock Learning equations for extrapolation and control.
\newblock \emph{CoRR}, abs/1806.07259, 2018.
\newblock URL \url{http://arxiv.org/abs/1806.07259}.

\bibitem[Schmidt and Lipson(2009{\natexlab{a}})]{Schmidt2009}
M.~Schmidt and H.~Lipson.
\newblock {D}istilling free-form natural laws from experimental data.
\newblock \emph{Science}, 324\penalty0 (5923):\penalty0 81--85,
  2009{\natexlab{a}}.

\bibitem[Schmidt and Lipson(2009{\natexlab{b}})]{schmidt09distilling}
Michael Schmidt and Hod Lipson.
\newblock Distilling free-form natural laws from experimental data.
\newblock \emph{Science}, 324\penalty0 (5923):\penalty0 81--85,
  2009{\natexlab{b}}.
\newblock \doi{10.1126/science.1165893}.
\newblock URL \url{https://www.science.org/doi/abs/10.1126/science.1165893}.

\bibitem[Srivastava et~al.(2014)Srivastava, Hinton, Krizhevsky, Sutskever, and
  Salakhutdinov]{srivastava14dropout}
Nitish Srivastava, Geoffrey Hinton, Alex Krizhevsky, Ilya Sutskever, and Ruslan
  Salakhutdinov.
\newblock Dropout: A simple way to prevent neural networks from overfitting.
\newblock \emph{Journal of Machine Learning Research}, 15\penalty0
  (56):\penalty0 1929--1958, 2014.
\newblock URL \url{http://jmlr.org/papers/v15/srivastava14a.html}.

\bibitem[Staelens et~al.(2013)Staelens, Deschrijver, Vladislavleva, Vermeulen,
  Dhaene, and Demeester]{Staelens2013}
Nicolas Staelens, Dirk Deschrijver, Ekaterina Vladislavleva, Brecht Vermeulen,
  Tom Dhaene, and Piet Demeester.
\newblock Constructing a no-reference h.264/avc bitstream-based video quality
  metric using genetic programming-based symbolic regression.
\newblock \emph{IEEE Trans. Cir. and Sys. for Video Technol.}, 23\penalty0
  (8):\penalty0 1322–1333, August 2013.
\newblock ISSN 1051-8215.
\newblock \doi{10.1109/TCSVT.2013.2243052}.

\bibitem[Uy et~al.(2011)Uy, Hoai, O'Neill, McKay, and
  Galvan-Lopez]{nguyen11semantically}
Nguyen~Quang Uy, Nguyen~Xuan Hoai, Michael O'Neill, R.~I. McKay, and Edgar
  Galvan-Lopez.
\newblock Semantically-based crossover in genetic programming: application to
  real-valued symbolic regression.
\newblock \emph{Genetic Programming and Evolvable Machines}, 12\penalty0
  (2):\penalty0 91--119, June 2011.
\newblock ISSN 1389-2576.
\newblock \doi{doi:10.1007/s10710-010-9121-2}.

\bibitem[Valipour et~al.(2021)Valipour, You, Panju, and
  Ghodsi]{valipour21symbolicgpt}
Mojtaba Valipour, Bowen You, Maysum Panju, and Ali Ghodsi.
\newblock Symbolicgpt: {A} generative transformer model for symbolic
  regression.
\newblock \emph{CoRR}, abs/2106.14131, 2021.
\newblock URL \url{https://arxiv.org/abs/2106.14131}.

\bibitem[Vaswani et~al.(2017)Vaswani, Shazeer, Parmar, Uszkoreit, Jones, Gomez,
  Kaiser, and Polosukhin]{vaswani17attention}
Ashish Vaswani, Noam Shazeer, Niki Parmar, Jakob Uszkoreit, Llion Jones,
  Aidan~N. Gomez, Lukasz Kaiser, and Illia Polosukhin.
\newblock Attention is all you need.
\newblock \emph{CoRR}, abs/1706.03762, 2017.
\newblock URL \url{http://arxiv.org/abs/1706.03762}.

\bibitem[Wadekar et~al.(2022)Wadekar, Thiele, Villaescusa-Navarro, Hill,
  Cranmer, Spergel, Battaglia, Anglés-Alcázar, Hernquist, and
  Ho]{wadekar22augmenting}
Digvijay Wadekar, Leander Thiele, Francisco Villaescusa-Navarro, J.~Colin Hill,
  Miles Cranmer, David~N. Spergel, Nicholas Battaglia, Daniel Anglés-Alcázar,
  Lars Hernquist, and Shirley Ho.
\newblock Augmenting astrophysical scaling relations with machine learning :
  application to reducing the sz flux-mass scatter, 2022.
\newblock URL \url{https://arxiv.org/abs/2201.01305}.

\bibitem[Werner et~al.(2021)Werner, Junginger, Hennig, and
  Martius]{werner21informed}
Matthias Werner, Andrej Junginger, Philipp Hennig, and Georg Martius.
\newblock Informed equation learning.
\newblock \emph{CoRR}, abs/2105.06331, 2021.
\newblock URL \url{https://arxiv.org/abs/2105.06331}.

\bibitem[Wilstrup and Kasak(2021)]{wilstrup21symbolic}
Casper Wilstrup and Jaan Kasak.
\newblock Symbolic regression outperforms other models for small data sets.
\newblock \emph{CoRR}, abs/2103.15147, 2021.
\newblock URL \url{https://arxiv.org/abs/2103.15147}.

\end{thebibliography}
}

\newpage
\phantomsection
\addcontentsline{toc}{section}{Supplementary Material}
\addtocontents{toc}{\protect\setcounter{tocdepth}{-5}}

\thispagestyle{empty}

\makeatletter
\vbox{%
\hsize\textwidth
\linewidth\hsize
\vskip 0.1in
\@toptitlebar
\centering
{\LARGE\bf Supplementary Material\par}
{\large \@title\par}
\@bottomtitlebar
\vskip 0.3in \@minus 0.1in
}
\makeatother

\appendix

In the supplementary material, we give more details on the experiments presented in the paper, data generator's and model's hyperparameters, and we also include additional experiments that evaluate the performance of the presented approach in more detail. In Section~\ref{sec:sampled_eq}, we investigate the effect of the number of sampled equations using Top-K sampling \citep{fan18hierarchical} on the model's performance and the inference time. In \Cref{sec:gen_examples}, we visualise several different model predictions using both univariate and bivariate \ours. In \Cref{sec:benchmark_functions}, we enumerate all benchmark functions that were used to compare our method with the current state of the art, and in \Cref{sec:benchmark_comp}, we present the exact values that we have measured while comparing the methods including the used hyperparameters. In \Cref{sec:hyper_dataset}, we present the exact hyperparameters that were used to generate both datasets, and in \Cref{sec:hyper_model}, we present the model hyperparameters, including its vocabulary and other training details.

Furthermore, we have also published the source code containing the necessary files to run the training and the inference \footnote{\url{https://github.com/vastlik/symformer}}. We have also included a video \footnote{\url{https://vastlik.github.io/symformer/video}} which shows a visualisation of inference on several functions. The process starts by sampling 256 equations and selecting the best eight functions from the sampled equations.
In the video, they are visualised as light blue functions. The orange points represent the sampled points used for the inference. 
We then show how the functions improve with each constant optimization step.

\section{Effect of number of sampled equations}
\label{sec:sampled_eq}
\begin{table}[htbp]
    \centering
        \caption{Performance of number of sampled equations for Top-K \citep{fan18hierarchical} with $K=20$ without a local gradient search on 256 equations. The time is in hours:minutes:seconds format.    \label{tab:sup_number_of_sampled_equations}\\} 
    \begin{tabular}{cccc}
        \# of equations & $R^2$ $\uparrow$ & RE $\downarrow$ & Time (hh:mm:ss) \\
        \cmidrule{1-2} \cmidrule{3-3} \cmidrule{4-4}
        16 & 0.998256 & 0.066528 & 00:04:46 \\
        32 & 0.998532 & 0.053106 & 00:04:36 \\
        64 & 0.999068 & 0.046484 & 00:06:46 \\
        128 & 0.999336 & 0.037968 & 00:07:19 \\
        256 & 0.999449 &  0.030616 & 00:19:04 \\
        512 & 0.999702 & 0.033389 & 00:30:05 \\
        1024 & 0.999826 & 0.022776 & 01:05:00 \\
        2048 & 0.999902 & 0.023793 & 01:03:57 \\
        4096 & 0.999942 & 0.017102 & 04:08:42 \\
        8192 & 0.999976 & 0.013216 & 06:15:53 \\
        16384 & 0.999985 & 0.008698 & 11:58:18

    \end{tabular}

\end{table}
First, we evaluate the model's performance with varying number of sampled equations during inference. In \Cref{tab:sup_number_of_sampled_equations}, we can see that the \ours's ability to predict the correct expression improves with the number of sampled equations, however, the time increases substantially, averaging approximately three minutes per equation for the largest number of equations. These results are measured by running the inference with Top-K sampling \citep{fan18hierarchical} with $K=20$ and without a local gradient search. If we had used a local gradient search, the $R^2$ and relative error (RE) would improve, however, the time would increase. 

\section{Examples of generated functions}
\label{sec:gen_examples}
\begin{figure}[htbp]
\definecolor{figblue}{HTML}{1f77b4}
\definecolor{figorange}{HTML}{d96e0f}
    \captionsetup[subfigure]{aboveskip=-1pt,belowskip=-1pt,font=footnotesize,justification=centering}
     \centering
     \begin{subfigure}[b]{0.48\textwidth}
         \centering
         \caption{{\textcolor{figblue}{GT}: $\cos{(0.77 (\tan{(x)})^{-1})}$,\\ \textcolor{figorange}{Pred}: $\cos(0.004 + 0.78 \cdot \cot{(x)})$}}
         \includegraphics[width=\textwidth]{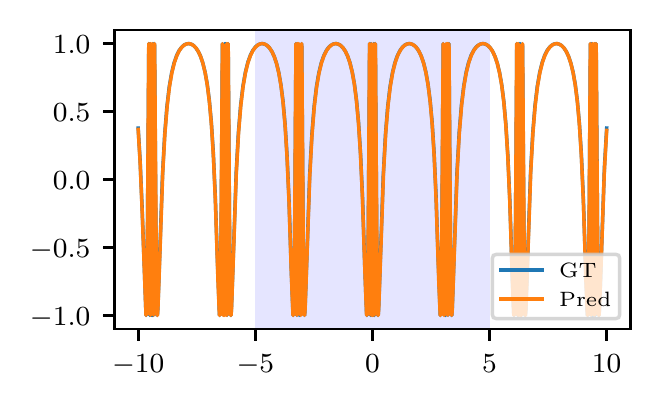}
     \end{subfigure}
     \begin{subfigure}[b]{0.48\textwidth}
         \centering
         \caption{{\textcolor{figblue}{GT}: $1.67 \cdot x \cdot \sin(-18.07+1.13x)$,\\ \textcolor{figorange}{Pred}: $-1.4\sqrt{(1.42 - 2 \cdot 10^{-4} \cdot x)} \cdot x \cdot \sin{(37-x)}$}}
         \includegraphics[width=\textwidth]{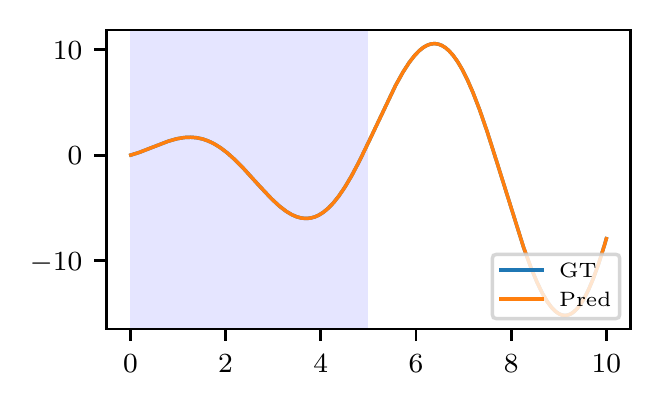}
    \end{subfigure}
    \begin{subfigure}[b]{0.48\textwidth}
         \centering
         \caption{{\textcolor{figblue}{GT}: $(\cos(x))^3 - (\sqrt{(x^{-1})})^{-1}$,\\ \textcolor{figorange}{Pred}: $(\cos(x))^3 - (\sqrt{(x^{-1})})^{-1}$}}
         \includegraphics[width=\textwidth]{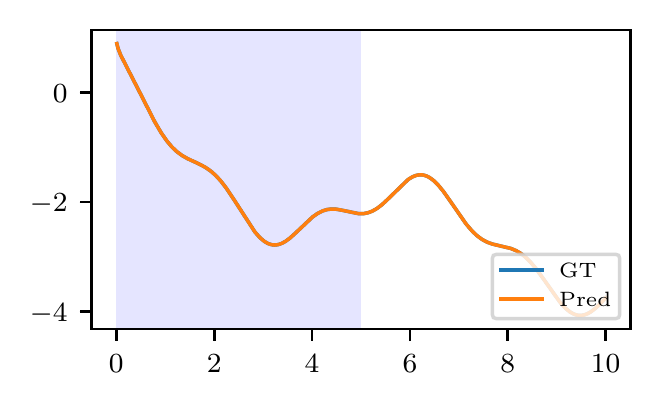}
    \end{subfigure}
    \begin{subfigure}[b]{0.48\textwidth}
         \centering
         \caption{\textcolor{figblue}{GT}: $x+\mathrm{acot}(0.28x)$,\\ \textcolor{figorange}{Pred}: $x+3.6x^{-1}\sqrt{(0.97+0.004x)}$}
         \includegraphics[width=\textwidth]{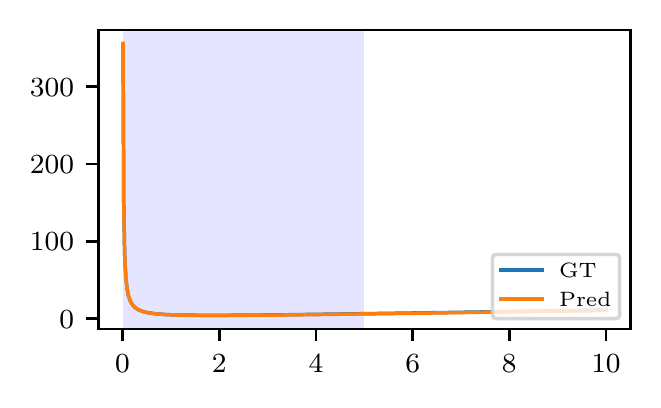}
    \end{subfigure}
    \begin{subfigure}[b]{0.48\textwidth}
         \centering
         \caption{\parbox{0.68\textwidth}{\textcolor{figblue}{GT}: $(\cos{x-\ln(x)})^2$,\\ \textcolor{figorange}{Pred}: $(\cos{x-\ln(x)})^2$}}
         \includegraphics[width=\textwidth]{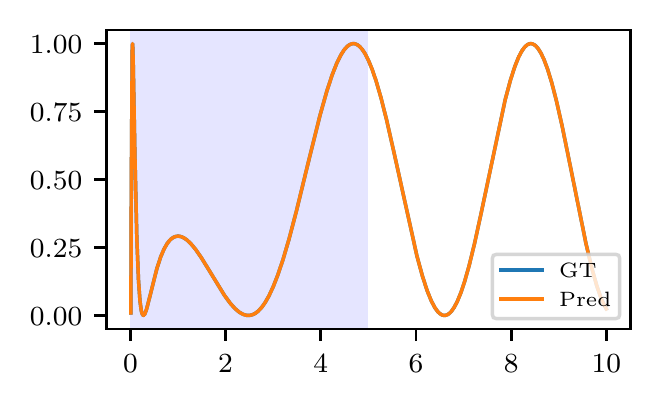}
    \end{subfigure}
    \begin{subfigure}[b]{0.48\textwidth}
         \centering
         \caption{\textcolor{figblue}{GT}: $0.27\cdot \mathrm{atan}{(2.25x^2)}$,\\ \textcolor{figorange}{Pred}: $\mathrm{atan}{(0.29 \cdot \mathrm{atan}{(2x^2))}}$}
         \includegraphics[width=\textwidth]{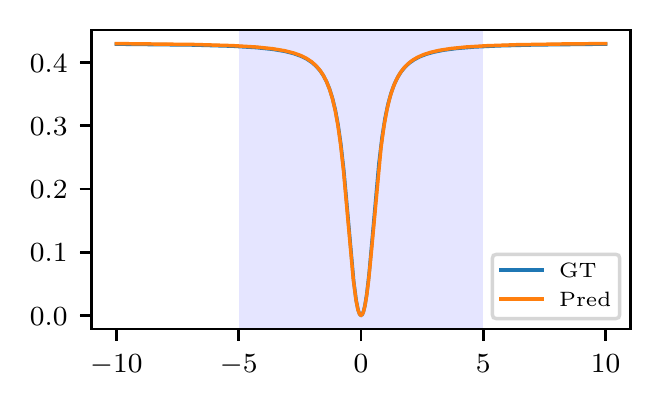}
    \end{subfigure}
    \caption{Examples of model predictions using Top-K sampling \citep{fan18hierarchical} with $K=20$ and 256 samples. The shaded area represents the sampling range. `\textcolor{figblue}{GT}' denotes ground-truth and `\textcolor{figorange}{Pred}', the univariate \Ours prediction.}
        \label{fig:supp_model_predictions}
\end{figure}

\begin{figure}[htbp]
\definecolor{figblue}{HTML}{1f77b4}
\definecolor{figorange}{HTML}{d96e0f}
    \captionsetup[subfigure]{aboveskip=-1pt,belowskip=-1pt,font=footnotesize,justification=centering}
     \centering
    \begin{subfigure}[b]{\textwidth}
         \centering
         \caption{\textcolor{figblue}{GT}: $-0.99-0.5y^2-0.5xy$, \quad\quad\quad\quad\quad \textcolor{figorange}{Pred}: $-1-0.5y^2-0.5xy$}
         \includegraphics[width=0.38\textwidth]{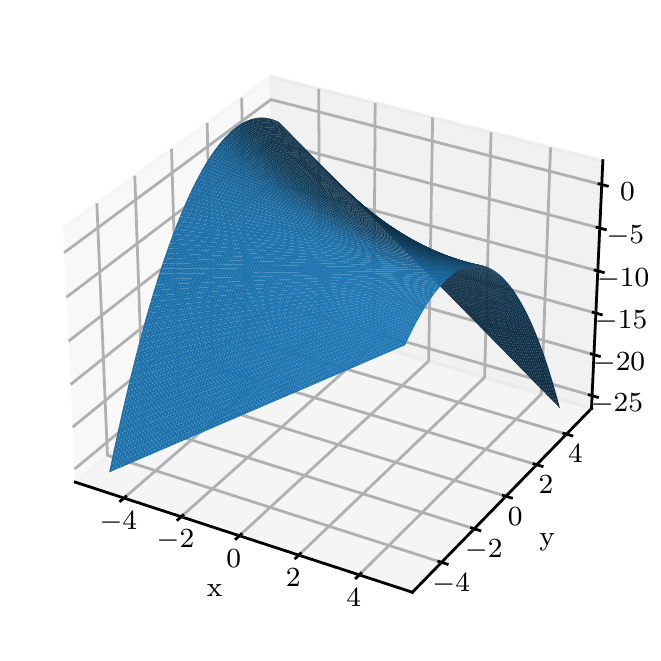}
         \includegraphics[width=0.38\textwidth]{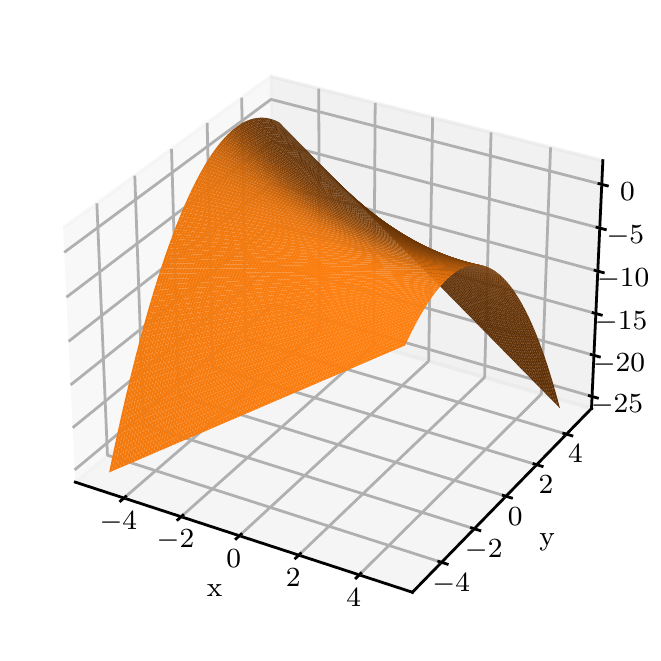}
    \end{subfigure}
    \begin{subfigure}[b]{\textwidth}
         \centering
         \caption{\textcolor{figblue}{GT}: $y^{0.0625}$, \quad\quad\quad\quad\quad\quad\quad\quad\quad \textcolor{figorange}{Pred}: $y^{0.0625}$}
         \includegraphics[width=0.38\textwidth]{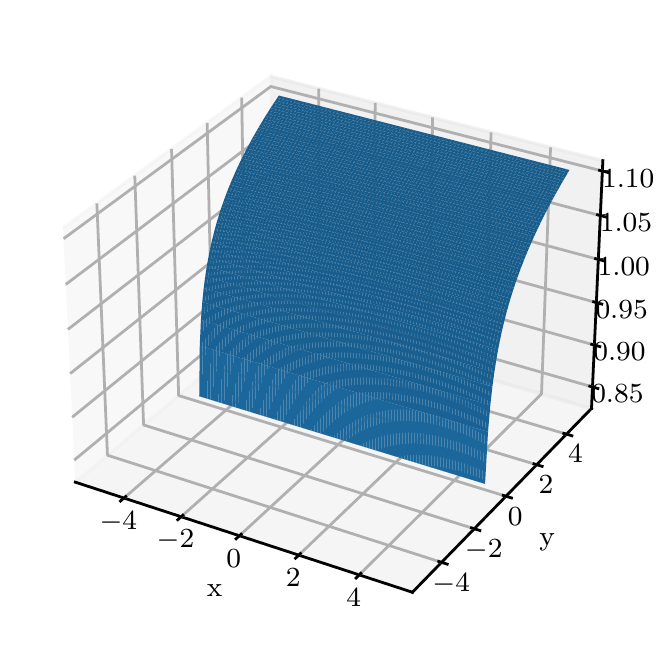}
         \includegraphics[width=0.38\textwidth]{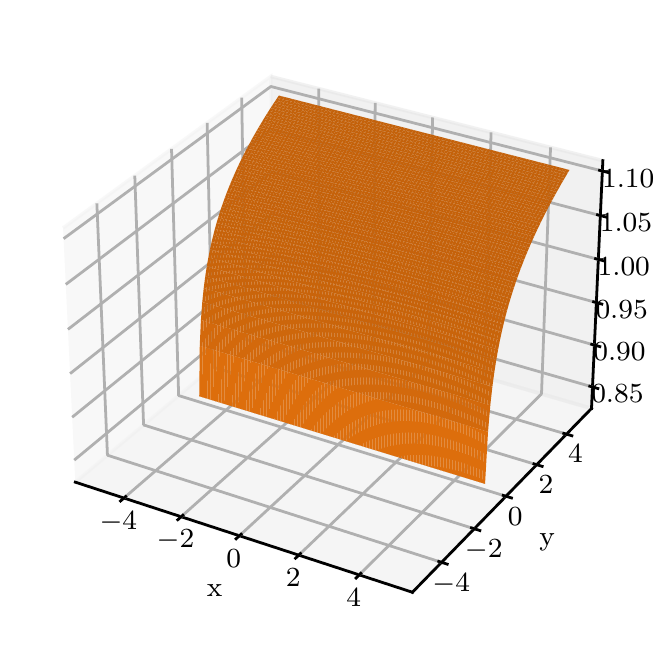}
    \end{subfigure}
    \begin{subfigure}[b]{\textwidth}
         \centering
         \caption{\textcolor{figblue}{GT}: $-1-0.94y+xy-y\sin{(x)}$, \quad\quad\quad\quad \textcolor{figorange}{Pred}: $-1.14-0.94y+xy-y\sin({x})$}
         \includegraphics[width=0.38\textwidth]{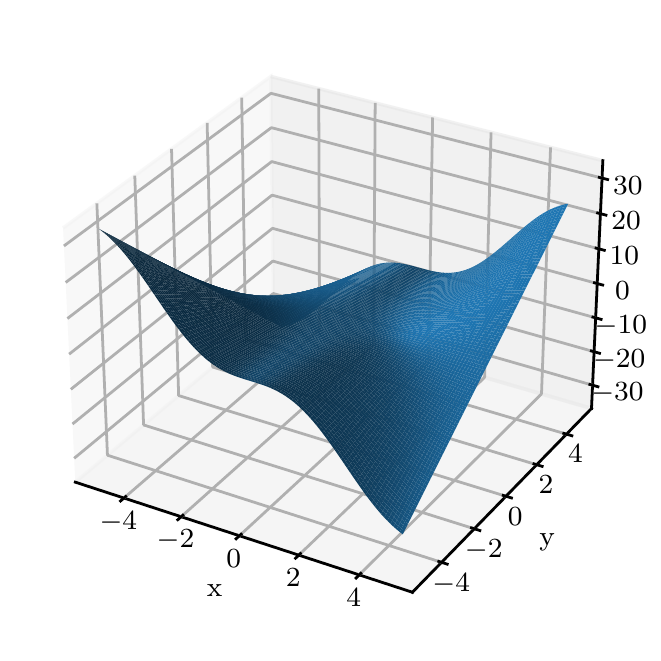}
         \includegraphics[width=0.38\textwidth]{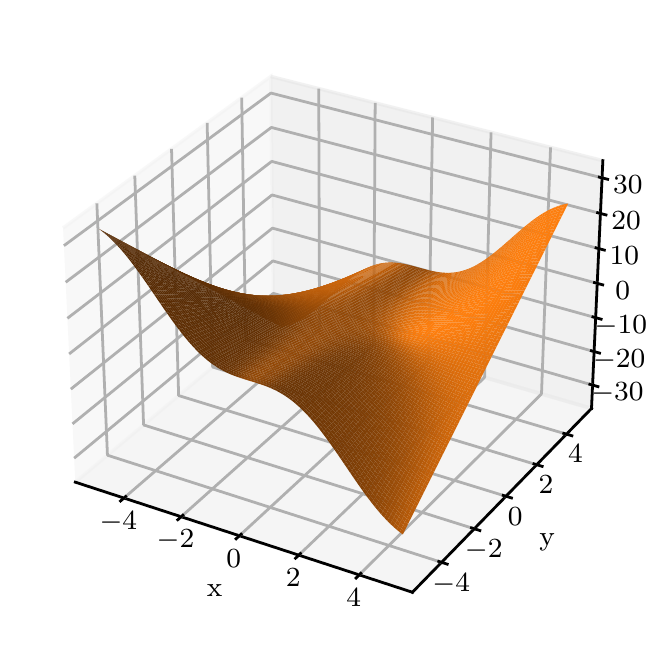}
    \end{subfigure}
    \caption{Examples of the bivariate \ours predictions using Top-K sampling \citep{fan18hierarchical} with $K=20$ and 256 samples. `\textcolor{figblue}{GT}' denotes ground-truth and `\textcolor{figorange}{Pred}', the model prediction.}
        \label{fig:supp_model_predictions_3d}
\end{figure}

We plotted several predictions generated by the \ours with the Top-K sampling \citep{fan18hierarchical} with $K=20$ with 256 sampled equations and the local gradient search. The shaded area represents the sampling range. The generated predictions can be seen in \Cref{fig:supp_model_predictions}. The predicted functions are not always the same as the original formula, however, they fit almost perfectly on the function domain. In \Cref{fig:supp_model_predictions_3d}, we show similar visualisation for the bivariate \ours.

\clearpage
\section{Benchmark functions}
\label{sec:benchmark_functions}
This section describes the exact functions used to compare the \ours with the current state-of-the-art methods. The benchmark's names and the contained functions can be seen in \Cref{tab:sup_benchmark}.

\begin{longtable}{lll}
\caption{Benchmark functions that we have used in our experiments. We have restricted ourselves only to the univariate and bivariate functions.\label{tab:sup_benchmark}}\\
    Name  & Expression \\
    \midrule
    Nguyen-1 & $x^3+x^2+x$ \\
    Nguyen-2 & $x^4+x^3+x^2+x$  \\
    Nguyen-3 & $x^5+x^4+x^3+x^2+x$  \\
    Nguyen-4 & $x^6+x^5+x^4+x^3+x^2+x$  \\
    Nguyen-5 & $\sin(x^2)\cos(x)-1$ \\
    Nguyen-6 & $\sin(x)+\sin(x+x^2)$  \\
    Nguyen-7 & $\ln(x+1)+\ln(x^2+1)$  \\
    Nguyen-8 & $\sqrt{x}$ \\
    Nguyen-9 & $\sin(x)+\sin(y^2)$ \\
    Nguyen-10 & $2\sin(x)\cos(y)$ \\
    Nguyen-11 & $x^y$  \\
    Nguyen-12 & $x^4-x^3+\frac{1}{2}y^2-y$ \\
    \midrule
    R-1    & $\frac{{\left(x+1\right)}^3}{{x}^2-x+1}$ \\
    R-2    & $\frac{{x}^5-3{x}^3+1}{{x}^2+1}$  \\
    R-3    & $\frac{{x}^6+{x}^5}{{x}^4+{x}^3+{x}^2+x+1}$  \\
    \midrule
    Livermore-1 & $\frac{1}{3}+x+\sin\left({x}^2\right)$ \\
    Livermore-2 & $\sin\left({x}^2\right) \cos\left(x\right)-2$ \\
    Livermore-3 & $\sin\left({x}^3\right) \cos\left({x}^2\right)-1$  \\
    Livermore-4 & $\ln(x+1)+\ln({x}^2+1)+\ln(x)$ \\
    Livermore-5 & ${x}^4-{x}^3+{x}^2-y$  \\
    Livermore-6 & $4{x}^4+3{x}^3+2{x}^2+x$ \\
    Livermore-7 & $\sinh(x)$  \\
    Livermore-8 & $\cosh(x)$ \\
    Livermore-9 & ${x}^9+{x}^8+{x}^7+{x}^6+{x}^5+{x}^4+{x}^3+{x}^2+x$  \\
    Livermore-10 & $6\sin\left(x\right) \cos\left(y\right)$ \\
    Livermore-11 & $\frac{{x}^2 {x}^2}{x+y}$  \\
    Livermore-12 & $\frac{{x}^5}{{y}^3}$ \\
    Livermore-13 & ${x}^{\frac{1}{3}}$ \\
    Livermore-14 & ${x}^3+{x}^2+x+\sin\left(x\right)+\sin\left({x}^2\right)$  \\
    Livermore-15 & ${x}^{\frac{1}{5}}$ \\
    Livermore-16 & ${x}^{\frac{2}{5}}$ \\
    Livermore-17 & $4\sin\left(x\right) \cos\left(y\right)$  \\
    Livermore-18 & $\sin\left({x}^2\right) \cos\left(x\right)-5$  \\
    Livermore-19 & ${x}^5+{x}^4+{x}^2+x$  \\
    Livermore-20 & $\operatorname{exp}\left({-x}^2\right)$ \\
    Livermore-21 & ${x}^8+{x}^7+{x}^6+{x}^5+{x}^4+{x}^3+{x}^2+x$ \\
    Livermore-22 & $\operatorname{exp}\left(-0.5{x}^2\right)$\\
    \midrule
    Koza-2 & $x^5-2x^3+x$  \\
    Koza-3 & $x^6-2x^4+x^2$ \\
    \midrule
    Keijzer-3 & $0.3 \cdot x \cdot \sin{(2 \cdot \pi \cdot x)}$  \\ 
    Keijzer-4 & $x^3\exp{(-x)}\cos{(x)} \sin{(x)} ({\sin{x}}^2\cos{x}-1)$ \\ 
    Keijzer-6 & $\frac{x*(x+1)}{2}$  \\ 
    Keijzer-7 & $\ln{x}$  \\ 
    Keijzer-8 & $\sqrt{x}$  \\ 
    Keijzer-9 & $\ln{(x+\sqrt{(x^2+1)})}$  \\ 
    Keijzer-10 & $x^y$   \\ 
    Keijzer-11 & $xy+\sin{((x-1)(y-1))}$   \\ 
    Keijzer-12 & $x^4-x^3+\frac{y^2}{2}-y$   \\ 
    Keijzer-13 & $6\sin{(x)}\cdot \cos{(y)}$ \\ 
    Keijzer-14 & $\frac{8}{2+x^2+y^2}$ \\ 
    Keijzer-15 & $\frac{x^3}{5}+\frac{y^3}{2}-y-x$  \\ 
    \midrule
    Constant-1 & $3.39x^3+2.12x^2+1.78x$  \\ 
    Constant-2 & $\sin{x^2}\cdot \cos{x}-0.75$ \\
    Constant-3 & $\sin{(1.5x)} \cdot \cos{(0.5y)})$ \\ 
    Constant-4 & $2.7x^y$  \\ 
    Constant-5 & $\sqrt{1.23x}$  \\ 
    Constant-6 & $x^{0.426}$ \\ 
    Constant-7 & $2\sin{(1.3x)} \cdot \cos{y}$ \\ 
    Constant-8 & $\ln(x+1.4)+\ln(x^2+1.3)$  \\ 

\\

\end{longtable}

\section{Benchmark comparison}
\label{sec:benchmark_comp}
This section presents a comparison for each equation from the used benchmarks. NSRS denotes Neural Symbolic Regression that Scales \citep{biggio21neural}, and DSO denotes Symbolic Regression via Neural-Guided Genetic Programming Population Seeding \citep{mundhenk21symbolic}. You can see that DSO tends to perform similarly to \ours, but it usually takes more time to find the underlying equation due to the usage of reinforcement learning, for example, in the case of the Kaijzer \citep{keijzer03improving} benchmark, where the DSO required thousands of seconds in most cases. For the NSRS, we can see that the times are consistent across the benchmarks, however larger than for \ours. The reason for this observation is that the model uses global optimisation to find the coefficients, which slows down the inference. The hyperparameters for the DSO can be found in \url{https://github.com/brendenpetersen/deep-symbolic-optimization/} and for the NSRS in \url{https://github.com/SymposiumOrganization/NeuralSymbolicRegressionThatScales}. The experiments were run using 32 CPU threads and 64 GB of RAM. From the results in \Cref{tab:sup_benchmark_results}, we can see that in most cases, the \ours is competitive in terms of $R^2$, however, it outperforms them in case of the time in the average case.

\begin{longtable}{ccccccc}
    
\caption{Comparison between methods on each of the benchmark functions, $R^2$ values are rounded to 4 decimals and time to whole seconds.\label{tab:sup_benchmark_results}}\\

    Name & \multicolumn{2}{c}{\ours} & \multicolumn{2}{c}{NSRS} & \multicolumn{2}{c}{DSO} \\
     & $R^2$ & Time (s) & $R^2$ & Time (s) & $R^2$ & Time (s) \\

    \midrule
    Nguyen-1 & 1    & 13 & 1 & 80 & 1 & 14 \\
    Nguyen-2 & 1    & 30 & 1 & 105 & 1 & 21 \\
    Nguyen-3 & 1    & 32 & 1 & 131 & 1 & 11 \\
    Nguyen-4 & 0.9998   & 136 & 0.5484 & 151 & 1 & 34 \\
    Nguyen-5 & 1    & 16 & 1 & 121 & 1 & 11 \\
    Nguyen-6 & 1    & 27  & 1 & 101 & 1 & 11 \\
    Nguyen-7 & 1    & 31 & 0.9967 & 145 & 1 & 134 \\
    Nguyen-8 & 1    & 5 & 1 & 122 & 1 & 53 \\
    Nguyen-9 & 1    & 13 & 1 & 166 & 1 & 11 \\
    Nguyen-10 & 1   & 14 & 1 & 190 & 1 & 23 \\
    Nguyen-11 & 1   & 8 & 0.999 & 296 & 1 & 11 \\
    Nguyen-12 & 1   & 57 & 1 & 503 & 0.9015 & 1004 \\
    \midrule
    R-1  & 1 & 18 & 1 & 135 & 0.9931 & 851 \\
    R-2  & 1 & 38 & 1 & 78 & 0.9681 & 778 \\
    R-3  & 0.9999 & 227 & 1 & 74 & 0.9790 & 937 \\
    \midrule
    Livermore-1 & 1 & 101 & 1 & 152 & 1 & 16 \\
    Livermore-2 & 1 & 15 & 0.1157 & 168 & 1 & 29 \\
    Livermore-3 & 1 & 15 & 0.1248 & 121 & 1 & 115   \\
    Livermore-4 & 1 & 22 & 0.9967 & 151 & 1 & 36 \\
    Livermore-5 & 1 & 54 & 1 & 463 & 1 & 167 \\
    Livermore-6 & 1 & 39 & 1 & 143 & 1 & 79\\
    Livermore-7 & 1 & 42 & 0.9999 & 138 & 0.9999 & 707 \\
    Livermore-8 & 0.9998 & 33 & 1 & 118 & 0.9999 & 771\\
    Livermore-9 & 0.9995 & 117 & 0.9831 & 130 & 1 & 872 \\
    Livermore-10 & 1 & 123 & 1 & 265 & 0.9694 & 971 \\
    Livermore-11 & 1 & 19 & 1 & 195 & 1 & 36 \\
    Livermore-12 & 1 & 14 & 1 & 212 & 1 & 68 \\
    Livermore-13 & 1 & 6 & 1 & 132 & 1 & 18 \\
    Livermore-14 & 1 & 30 & 1 & 572 & 1 & 110 \\
    Livermore-15 & 1 & 70 & 1 & 154 & 1 & 146 \\
    Livermore-16 & 1 & 33 & 0.9986 & 126 & 1 & 286  \\
    Livermore-17 & 1 & 13 & 1 & 288 & 0.9972 & 233 \\
    Livermore-18 & 1 & 15 & 0.2679 & 183 & 0.9677 & 891\\
    Livermore-18 & 1 & 32 & 1 & 150 & 1 & 30\\
    Livermore-20 & 1 & 7 & 1 & 115 & 1 & 14\\
    Livermore-21 & 0.9999 & 138 & 0.9944 & 148 & 1 & 120\\
    Livermore-22 & 1 & 8 & 1 & 124 & 0.9992 & 364\\
    \midrule
    Koza-2 & 1 & 20 & 1 & 73 & 1 & 27 \\
    Koza-3 & 1 & 182 & 1 & 150 & 1 & 408 \\
    \midrule
    Keijzer-3 & 1 & 49 & 0.7549 & 174 & 0.6454 & 5802 \\ 
    Keijzer-4 & 0.9887 & 58 & 0.9991 & 179 & 0.8990 & 9171 \\ 
    Keijzer-6 & 1 & 13 & 1 & 108 & 1 & 510 \\ 
    Keijzer-7 & 1 & 5 & 1 & 138 & 1 & 1678 \\ 
    Keijzer-8 & 1 & 5 & 1 & 203 & 1 & 86 \\ 
    Keijzer-9 & 1 & 116 & 0.996 & 126 & 1 & 1304 \\ 
    Keijzer-10 & 1 & 12 & 0.9335 & 316 & 0.9856 & 2926 \\ 
    Keijzer-11 & 1 & 78 & 1 & 240 & 0.9536 & 5235 \\ 
    Keijzer-12 & 1 & 48 & 1 & 521 & 1 & 8124 \\ 
    Keijzer-13 & 1 & 96 & 1 & 321 & 0.9526 & 6848 \\ 
    Keijzer-14 & 1 & 37 & 1 & 274 & 1 & 3864 \\ 
    Keijzer-15 & 1 & 67 & 1 & 266 & 0.9999 & 1601\\ 
    \midrule
    Constant-1 & 1 & 23 & 1 & 175 & 1 & 972 \\ 
    Constant-2 & 1 & 162 & 0.0996 & 130 & 1 & 4836 \\
    Constant-3 & 1 & 89 & 1 & 348 & 1 & 976\\ 
    Constant-4 & 1 & 11 & 0.9997 & 317 & 1 & 707 \\    
    Constant-5 & 1 & 57 & 1 & 127 & 1 &  1094 \\ 
    Constant-6 & 1 & 19 & 1 & 221 & 1  & 907\\ 
    Constant-7 & 1 & 146 & 1 & 353 & 1 & 4186\\ 
    Constant-8 & 1 & 220 & 1 & 172 & 1 & 8851 \\ 

\end{longtable}

\section{Hyperparameters for dataset generation}
\label{sec:hyper_dataset}
In this section, we describe the exact hyperparameters used to generate the bivariate and univariate datasets. Note that in the case of the bivariate dataset, the probability of selecting the x variable was the same as the y variable. The probability of selecting the unary operation is also the same as the binary operation. The unnormalized probabilities for unary operations can be seen in \Cref{table:unary-probs}, for binary in \Cref{table:binary-probs} and for leafs in \Cref{table:leaf-values}.

\begin{table}[htbp]
\centering
\caption{Unnormalised probabilities of unary operators used by the dataset generator. We have also used special operations to generate them more often.\label{table:unary-probs}\\}

\begin{tabular}{lll}
Operation & Mathematical meaning & Unnormalized probability \\ 
\cmidrule{1-3}

pow2 & $(\cdot)^2$ & 8 \\
pow3 & $(\cdot)^3$ & 6 \\  
pow4 & $(\cdot)^4$ & 4 \\ 
pow5 & $(\cdot)^5$ & 4 \\ 
pow6 & $(\cdot)^6$ & 3 \\ 
inv & $(\cdot)^{-1}$ & 8 \\
sqrt & $\sqrt{\cdot}$ & 8 \\ 
exp & $\exp{\cdot}$ & 2 \\ 
ln & $\ln{\cdot}$ & 4 \\ 
sin & $\sin{\cdot}$ & 4 \\ 
cos & $\cos{\cdot}$ & 4 \\ 
tan & $\tan{\cdot}$ & 2 \\ 
cot & $\cot{\cdot}$ & 2 \\ 
asin & $\arcsin{\cdot}$ & 1 \\  
acos & $\arccos{\cdot}$ & 1 \\
atan & $\arctan{\cdot}$ & 1 \\ 
acot & $\text{arccot}{\cdot}$ & 1 \\ 
\end{tabular}
\end{table}

\begin{table}[t]
\centering
\caption{Unnormalised probabilities of binary operators as used by the dataset generator.\hspace{\textwidth}\label{table:binary-probs}\\}

\begin{tabular}{ll}
Operation & Unnormalized probability \\ \cmidrule{1-2}
+ & 8 \\ 
- & 5 \\  
* & 8 \\  
/ & 5 \\  
pow & 2 \\ 
\end{tabular}
\end{table}

\begin{table}[t]
\centering
\caption{Unnormalised probabilities of leaf values as used by the dataset generator.\hspace{\textwidth}\label{table:leaf-values}}

\begin{tabular}{ll}
Operation & Unnormalized probability \\ \cmidrule{1-2}

variable & 20 \\ 
integer in interval $[-5,5]$ (excluding zero) & 10 \\  
float in interval $[-5,5] $ & 10 \\ 
zero & 1 \\ 
\end{tabular}
\end{table}
\clearpage
\section{Model hyperparameters and vocabulary}
\label{sec:hyper_model}
In this section, we give more details about the model's hyperparameters that were used to train our model. We have generated two datasets -- one with univariate functions containing 130 million equations and the second one with bivariate functions containing 100 million equations. Note that the bivariate dataset can also contain functions that have only one input variable. The main difference between these two datasets is the number of sampled points, 100 for the univariate case, and 200 for the bivariate case. For the model, we have decided to use a large encoder which has the majority of parameters (approximately 75 millions) and the rest for the decoder. The idea is that the encoder should not just encode the points, but also represent the function on a high level such that the decoder only prints the representation as a sequence of symbols. The full set of hyperparametrs can be seen in \Cref{tab:sup_hyperparameters}.

We also present our model's vocabulary which can be seen in \Cref{tab:sup_vocab} where we have used the common functions such as $\ln$ or $\sin$. However, we have also used some operators, that are not elementary functions, \eg, $\text{neg}(x)=-1x$. We have also added integers from $-5$ to 5, so it is easier for the model to represent them. We also did not use the hyperbolic trigonometric functions, since they can be represented by an equivalent expression e.g. $\sinh{x}=\frac{\exp{(x)}-\exp{(-x)}}{2}$.

\begin{table}[htbp]
\centering
\caption{Hyperparameters for both the univariate and bivariate models.\label{tab:sup_hyperparameters}\\}

\begin{tabular}{ll}
\midrule

\multicolumn{2}{c}{Encoder}    \\
\multicolumn{1}{l}{Number of row wise FF} & 2 \\ 
\multicolumn{1}{l}{Number of heads} & 12 \\
\multicolumn{1}{l}{Number of layers} & 4 \\
\multicolumn{1}{l}{Dimension of model} & 384 \\
\multicolumn{1}{l}{Dimension of the first FF layer} & 1536 \\
\multicolumn{1}{l}{Dimension of the second FF layer} & 384 \\
\multicolumn{1}{l}{Number of inducing points} & 64 \\
\multicolumn{1}{l}{Number of seed vectors} & 32 \\
\multicolumn{1}{l}{Dropout rate} & 0.1 \\

\midrule

\multicolumn{2}{c}{Decoder} \\
\multicolumn{1}{l}{Dimension of model} & 512 \\
\multicolumn{1}{l}{Number of heads} & 12 \\
\multicolumn{1}{l}{Dimension of FF layer} & 2048 \\
\multicolumn{1}{l}{Number of layers} & 4 \\
\multicolumn{1}{l}{Dropout rate} & 0.1 \\
\multicolumn{1}{l}{Vocabulary size} & 54 \\

\midrule

\multicolumn{2}{c}{Dataset} \\

\multicolumn{1}{l}{Number of equations (bivariate)} & 130 million (100 million) \\
\multicolumn{1}{l}{Number of sampled points (bivariate)} & 100 (200)\\

\midrule

\multicolumn{2}{c}{Training} \\
\multicolumn{1}{l}{Number of epochs (bivariate)} & 390 (300)\\
\multicolumn{1}{l}{Starting $\sigma^2$ noise} & 0.1 \\
\multicolumn{1}{l}{Ending regression $\lambda$} & 1 \\

\end{tabular}
\end{table}

\begin{table}[htbp]
\centering
\caption{Vocabulary as used by the model. Besides the elementary functions, it contains special functions such as pow2 and pow3, which are commonly used.\label{tab:sup_vocab}\\}

\begin{tabular}{ll}
Token & Mathematical meaning \\  \cmidrule{1-2}
Integers from [-5, 5] & Integers from [-5, 5] \\ 
Variables & Variables \\  
pow & $(\cdot)^\cdot$ \\ 
+ & addition \\ 
* & multiplication \\ 
sqrt & $\sqrt{\cdot}$ \\ 
pow2 & $(\cdot)^2$ \\ 
pow3 & $(\cdot)^3$ \\
ln & $\ln{\cdot}$ \\ 
exp & $\exp{\cdot}$ \\ 
sin & $\sin{\cdot}$ \\ 
cos & $\cos{\cdot}$ \\ 
tan & $\tan{\cdot}$\\ 
cot & $\cot{\cdot}$ \\ 
asin & $\arcsin{\cdot}$ \\ 
acos & $\arccos{\cdot}$ \\
atan & $\arctan{\cdot}$ \\
acot & $\text{acot}\cdot$ \\ 
neg & $(-1)(\cdot)$ \\ 
C & Constants, optionally C-10, \ldots, C10 for scientific-like encoding \\  

\end{tabular}
\end{table}

\end{document}